%% file: main.tex
\documentclass[final]{cvpr}

\usepackage{times}
\usepackage{epsfig}
\usepackage{graphicx}
\usepackage{amsmath}
\usepackage{amssymb}
\usepackage{algorithm}
\usepackage{algorithmic}
\usepackage[english]{babel}

\usepackage{cuted}
\usepackage{capt-of}
\usepackage{mathtools}

\usepackage{tabularx}
\usepackage{booktabs}
\usepackage[font=small]{caption}

\usepackage[pagebackref=true,breaklinks=true,colorlinks,bookmarks=false]{hyperref}

\newif\ifcomments
\commentstrue 
\ifcomments
    \newcommand{\david}[1]{\textcolor{red}{[DJ: #1]}}
    
    \newcommand{\daniel}[1]{\textcolor{green}{[DL: #1]}}
    \newcommand{\jiaye}[1]{\textcolor{magenta}{[J: #1]}}
    \newcommand{\soumyadip}[1]{\textcolor{blue}{[SS: #1]}}
\else
    \providecommand{\david}[1]{}
    \providecommand{\daniel}[1]{}
    \providecommand{\jiaye}[1]{}
    \providecommand{\soumyadip}[1]{}
\fi

\def\cvprPaperID{10304} 
\def\confYear{CVPR 2021}

\begin{document}
\title{Shape and Material Capture at Home}

\author{Daniel Lichy$^1$ \hspace{30pt} Jiaye Wu$^1$ \hspace{30pt} Soumyadip Sengupta$^2$ \hspace{30pt} David W. Jacobs$^1$ \\
$^1$University of Maryland, College Park \hspace{30pt} $^2$University of Washington\\
{\tt\small dlichy@umd.edu, jiayewu@umiacs.umd.edu, soumya91@cs.washington.edu, djacobs@cs.umd.edu}
}

\maketitle


\input{abstract}
\input{intro.tex}
\input{prior_work.tex}
\input{model.tex}
\input{compare_sota}
\input{ablation}
\input{conclusion}

{\small
\bibliographystyle{ieee_fullname}
\bibliography{egbib}
}

\newpage
\input{appendix}

\end{document}

%% file: abstract.tex
\begin{abstract}
\vspace{-1em}
In this paper, we present a technique for estimating the geometry and reflectance of objects using only a camera, flashlight, and optionally a tripod. We propose a simple data capture technique in which the user goes around the object, illuminating it with a flashlight and capturing only a few images.Our main technical contribution is the introduction of a recursive neural architecture, which can predict geometry and reflectance at $2^{k}\times2^{k}$ resolution given an input image at $2^{k}\times2^{k}$ and estimated geometry and reflectance from the previous step at $2^{k-1}\times2^{k-1}$. This recursive architecture, termed RecNet, is trained with 256$\times$256 resolution but can easily operate on 1024$\times$1024 images during inference. 
We show that our method produces more accurate surface normal and albedo, especially in regions of specular highlights and cast shadows, compared to previous approaches, given three or fewer input images. Our model and code is available at {\small \url{https://dlichy.github.io/ShapeAndMaterialAtHome/}}.
\end{abstract}

%% file: intro.tex
\vspace{-1em}
\section{Introduction}
\vspace{-0.5em}
Capturing an object's shape and material is a long-studied problem in Computer Vision and Graphics, with a broad range of applications such as Augmented and Virtual Reality. With the rise of e-commerce, it is now even more important to develop a system that allows sellers to capture 3D shape and material of their product with relative ease. It can also be extremely useful for digital artists who can use captured 3D objects as a starting point for their models.

In this work, we present a system where the shape and reflectance of an object can be captured by a user with everyday household items. We propose a capture setup that requires a camera, flashlight, and ideally a tripod. The user can capture one or multiple images of an object by illuminating the object with a flashlight from multiple directions. This relatively straight-forward data capture setup provides us with important photometric cues necessary for high-quality reconstruction.

High quality object geometry can be recovered using a large number of images either from different views (Multi-View Stereo) \cite{mildenhall2020nerf,yao2018mvsnet,yao2019recurrent}, or from different lighting variations (Photometric Stereo) \cite{chen2019SDPS_Net,Chen_2018_ECCV,chen2020learned} . Our work falls in the category of both Photometric Stereo (PS) and Shape-from-Shading (SfS). PS techniques generally require a light-stage setup in a dark room to capture an object. Although these techniques often produce high-quality reconstruction given a large enough number of images, it is extremely difficult to create such a capture setup at home. On the other hand, single image based methods have shown success in capturing geometry and material using a camera with co-directional flash \cite{li2018learning}, often with an extra image without the flash \cite{Boss2020-TwoShotShapeAndBrdf}. However, these approaches rely on priors learned from synthetic data with limited generalization to real data, especially with slight variation in capture setup. In this work, we aim to meet reconstruction quality and ease of capture in the middle. We propose to capture only a few images (at most six) of the object from approximately known lighting directions in a weakly calibrated fashion to achieve satisfactory reconstruction quality.

We introduce a novel, recursive architecture, RecNet, that is capable of predicting normal, albedo and roughness at high resolution. This allows our approach to create a high-quality reconstruction with fine details.  RecNet is trained to predict normal, albedo and roughness at $2^{k}\times2^{k}$ resolution given an input image at $2^{k}\times2^{k}$ and predicted normal, albedo and roughness from the a previous iteration at $2^{k-1}\times2^{k-1}$.  We parameterize shape with surface normal and reflectance with albedo and roughness following the Cook-Torrance model.  RecNet is trained with resolutions between 64$\times$64 and 256$\times$256, but can then be applied to images of arbitrary resolution.  To start this recursive process off, we use a small network, InitNet, that takes in the image at 32$\times$32 and estimates the normals, albedo, and roughness at the same resolution.  We demonstrate that RecNet produces high-quality results for 1024$\times$1024 images, yielding a mesh with enough vertices to capture fine detail.

An alternative approach is to train at the same resolution as inference.  However, training on higher resolution input offers its own unique challenges; it requires high-quality synthetic data and a very deep convolutional network. This would necessitate large amounts of GPU memory and training time, making it extremely difficult if not impossible. Thus PS techniques that require a large number of images (50-100) often train on pixels or patches \cite{ taniai2018neural,ikehata2018cnn,logothetis2020px, chen2019SDPS_Net,Chen_2018_ECCV,chen2020learned}, but they do not perform well when using a small number of images. Most previous single image techniques can only be run at $256 \times 256$ \cite{li2018learning,Boss2020-TwoShotShapeAndBrdf}. This is because photometric cues from a single image are not sufficient to determine normals using local information alone, and the global priors the network learns from 256$\times$256 data do not transfer well to higher resolution. A network trained on lower resolution, when applied to higher resolutions effectively shrinks its receptive field, implicitly assuming that the normals are conditionally independent of the rest of the image, preventing it from capturing global context. In contrast, we assume that normals at higher resolution depend on the normals at lower resolution within the receptive field via our recursive architecture, RecNet. Normals at lower resolution reflect larger global context with the same receptive field and thus help our network to reconstruct better shape and reflectance utilizing the global context.

Our network can also handle anywhere from one to six images. When multiple images are used, we compare with a current state-of-the-art PS technique, SDPS-Net \cite{chen2019SDPS_Net}, and when a single image is used as input, we compare with \cite{li2018learning,Boss2020-TwoShotShapeAndBrdf}, which assume flash-light co-directional with the camera. We predict higher quality surface normals compared to SDPS-Net, especially on objects with spatially varying BRDF and when given a low number of input images. While SDPS-Net only predicts geometry, we also predict material reflectance as albedo and roughness. In comparison to single image techniques, we predict better surface normals and albedo, especially in the regions of cast shadows and specular reflectance.

In summary, our contributions are as follows: 

$\bullet$ We introduce a weakly calibrated photometric stereo technique, where shape and material of an object can be captured easily at home with a camera, flashlight, and optionally a tripod. 
$\bullet$ We present a recursive multi-resolution architecture that can handle a varying number of input images (1-6) at arbitrary resolution, which can be higher than the training resolution (training at 256, inference at 1024) with minimal artifacts.

%% file: prior_work.tex
\section{Prior Work}

Photometric Stereo (PS), first introduced in \cite{woodham1980photometric}, aims to reconstruct the shape of an object from images under varying illumination, and is a long studied problem in computer vision (see \cite{ackermann2015survey,durou2020advances,diligent_data} for surveys). When intensity and direction of the lights are unknown the problem is referred to as Uncalibrated PS (UPS) \cite{basri2007photometric}, which is often more challenging due to ambiguity \cite{belhumeur1999bas}. Previous approaches either assume Lambertian reflectance \cite{favaro2012closed,queau2015solving,haefner2019photometric} or aim to model non-diffuse reflectance using various BRDF models \cite{mecca2016unifying,georghiades2003incorporating,okabe2009attached,lu2017symps}. A few works \cite{haefner2019variational,mo2018uncalibrated,brahimi2020well,jung2015one} aim to solve PS for more general environment lighting.

\begin{figure*}[h!]
\centering
    \includegraphics[width=0.98\linewidth]{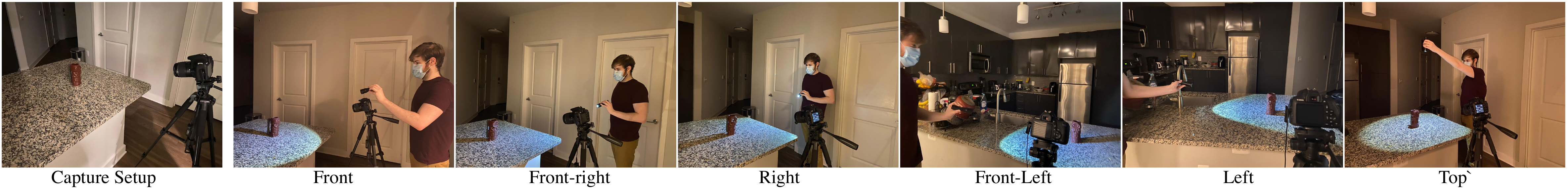}
    \vspace{-0.5em}
   \caption{Our at-home capture setup uses a camera, a flashlight, and optionally a tripod and a remote. A user moves around the object to capture a single or multiple images with variation in lighting. Images are captured from the `Front' and the five positions shown above.}
\label{fig:light_dirs}
\vspace{-1.5em}
\end{figure*}

With the recent success of deep learning, researchers have shown renewed interest in pushing the boundaries of PS. Recent works \cite{santo2017deep,ikehata2018cnn,logothetis2020px} focus on the calibrated setting with a large number of input images and learn on a per-pixel basis, thus ignoring global context. Taniai \textit{et al.} \cite{taniai2018neural} proposed a network that is trained per object to predict normal and BRDF using unsupervised reconstruction loss with known lighting. Chen \textit{et al.} \cite{chen2018ps} trained a network on synthetic data to predict normals with known lighting. For UPS, recent works focus on predicting the lighting and then using it to predict normals \cite{chen2019self}, often in a recursive fashion \cite{chen2020learned}, and can handle spatially varying BRDF \cite{chen2020deep}. While most PS methods require a large number of images (50-100), researchers have also attempted to solve PS with few images \cite{sengupta2018solving,mecca2014direct,queau2017photometric,li2019learning,zheng2019spline}. Researchers have also proposed techniques to solve PS with known direction and unknown intensity \cite{cho2018semi,queau2017semi,queau2017photometric}, and with inaccurate direction and intensity \cite{queau2017non}. In this work, our system assumes that the lighting direction is known to within $10^{\circ}$ to $15^{\circ}$, and the intensity is unknown. It requires only a few images (1-6) and predicts both normal and spatially-varying BRDF.

While PS usually requires a large number of observations, another research direction, Shape from Shading \cite{zhang1999shape}, attempts to predict shape and reflectance from a single image. Previous approaches like Barron and Malik \cite{barron2014shape} relied on extensive, manually designed shape priors. With recent advances in deep networks such priors are often learned from large volumes of synthetic data. Recent works made progress in solving Shape from Shading for faces \cite{kemelmacher20103d,sengupta2018sfsnet,Lattas_2020_CVPR} and Inverse Rendering for general scenes \cite{yu2019inverserendernet,neuralSengupta19,li2020inverse}, all from a single image. For a generic object, researchers have attempted to estimate reflectance and illumination from a single image \cite{LN2012,LN2016,georgoulis2016delight,georgoulis2017reflectance,rematas2016deep,LIMEMeka:2018}. Liu \textit{et al.} \cite{liu2017material} predicted normal and DS-BRDF from a single image by training on ShapeNet objects. Li \textit{et al.} \cite{li2018learning} proposed a framework where a single image of an object is captured with a flash co-directional with the camera and predicted normal, depth, spatially varying BRDF and illumination. They trained a cascaded architecture on synthetic data, generated similarly to our approach. An extension of single image based shape from shading is to capture two images, with and without a co-direction flash, as presented in Boss \textit{et al..} \cite{Boss2020-TwoShotShapeAndBrdf}, which also uses a cascaded architecture. Both \cite{li2018learning,Boss2020-TwoShotShapeAndBrdf} can only make predictions at 256$\times$256 resolution. In contrast, we present a unified architecture that predicts state-of-the-art normal and spatially varying BRDFs from a single image or from a few images (2-6) illuminated with a flashlight and ambient light (in at least one image the flashlight is approximately co-directional with the camera), at 1024$\times$1024 resolution. 

A key contribution of our technique is a recursive multi-scale network, where a single network is trained with multiple resolutions. The importance of multi-scale information in normal estimation from a single image goes back well before the age of deep learning. This is due to ambiguities that cannot be resolved with local information alone but require global context. Separate networks for local and global context and a fusion step to exploit multi-scale information were used in \cite{multi-scale-normal-gupta,multi-scale-normal-Fergus}. Recent deep networks often aim to achieve this by using a cascaded architecture where the later stages in the cascade are responsible for learning details \cite{li2018learning,li2020inverse}, but all stages operate at the same resolution. However, these methods trained for 256$\times$256 cannot operate on 1024$\times$1024, as they involve fully connected layers. Often optimization is also used for refining the network prediction \cite{bi2020deep} at the cost of runtime. Researchers have also explored cascaded architectures that operate at different resolutions in segmentation \cite{lin2019refinenet,Lin:2017:RefineNet}, super-resolution \cite{deeplaplacianPyramid}, and generative models \cite{chen2017photographic,karras2017progressive}. In contrast to these methods, our network uses the same weights at each step for predicting normal, albedo and roughness at resolution $2^{k}\times2^{k}$ given an input image at $2^{k}\times2^{k}$ and predicted normal, albedo and roughness at $2^{k-1}\times2^{k-1}$. This also enables us to operate on images larger than those we train on.

%% file: model.tex
\begin{figure*}[h!]
\centering
    \includegraphics[width=0.98\linewidth]{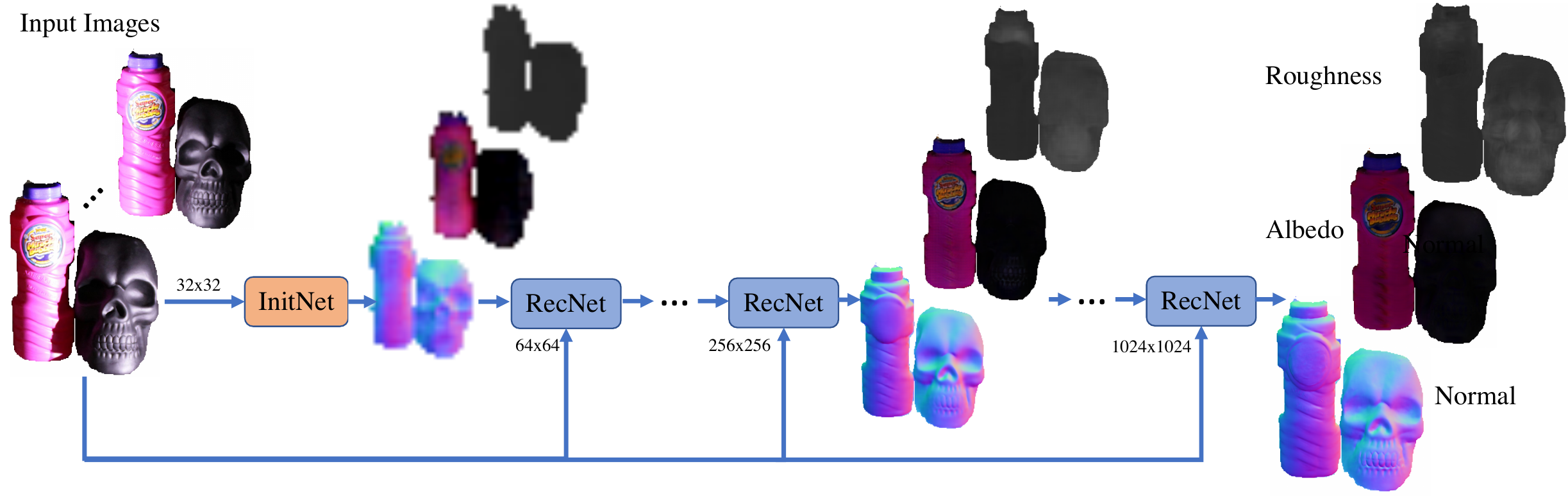}
    \vspace{-0.5em}
   \caption{We propose a recursive multi-resolution architecture, RecNet, that predicts surface normal, albedo and roughness from the input image(s) and from the prediction at the previous step by continuously upsampling by a factor of 2. The recursion is initialized by InitNet.}
\label{fig:network}
\vspace{-1.5em}
\end{figure*}

\section{Our Approach}
Our goal is to capture high-quality shape and reflectance of household objects with minimal effort. To achieve this goal, we make several design choices. First, we parameterize shape with surface normals, which will then be integrated into a depth map and converted into a mesh representation. Second, we parameterize spatially-varying BRDF (SV-BRDF) with albedo and roughness using the Cook-Torrance BRDF model \cite{cook-torrance} (for details please see Appendix \ref{brdf_model}). We aim to reconstruct surface normal, albedo and roughness at 1024$\times$1024 resolution, in contrast to 256$\times$256 used in \cite{Boss2020-TwoShotShapeAndBrdf,li2018learning}, such that the reconstructed mesh can preserve details. Finally, our capture setup includes a camera mounted on an inexpensive tripod about 2' from the object. Then the user captures multiple images of the object by walking around with a flashlight, with at least one image where the light is roughly co-directional with the camera. Our data capture setup inspires us to generate synthetic data reflective of our real data.

\subsection{Our Capture Setup}
\label{sec:real_dataset}

We want our capture setup to be simple such that, with instructions, a user can easily replicate it in their home, and also provide enough distinct observations for high-quality reconstruction. For our capture setup we used the following items (i) A CANON 2000D DSLR camera; (ii) a tripod; (iii) a remote trigger; and (iv) an LED flashlight. As we show in Figure \ref{fig:teaser} our capture setup can be performed with just a camera phone, stand, and flashlight.

Our capture protocol is as follows. The object is placed on a flat surface in a room dimly lit relative to the flashlight (\textit{dark room not required}). The camera is pointed toward the object; its optical axis making approximately a 20 to 40$^\circ$ angle with the surface. The user can then take one to six images by moving around with a flashlight pointed at the object. In one image, the light should be approximately co-directional with the camera. For six images, the flashlight is pointed at the object from the following directions: right ($+90^{\circ}$), front-right ($+45^{\circ}$), co-directional with the camera ($0^{\circ}$), front-left ($-45^{\circ}$), left ($-90^{\circ}$), and directly above. Note that these directions are weakly calibrated, i.e. our system roughly knows the location of the light and can tolerate $\pm15^{\circ}$ error in angular direction. We chose these directions because we felt they are easy for the user to estimate. Please see Fig \ref{fig:light_dirs}, which depicts a user capturing an object following our capture setup. To evaluate our method, we capture a large real dataset of 111 scenes containing a variety of objects we obtained at various discount stores.

\subsection{Synthetic Data}

Since it is extremely difficult to capture real-world objects with ground-truth shape and reflectance, we rely on synthetic data to train our network. Our synthetic data generation is motivated by our real-world data capture setup. We perform synthetic data generation with primitive shapes, similar to Xu \textit{et. al} \cite{xu2018deep}. We consider 5,000 objects, each consisting of 1-9 primitive shapes (cubes, ellipsoids, and cylinders) with surface perturbations of various frequencies. Additionally, we consider 14 realistic shapes from the Sculptures dataset \cite{sculpture_data}.

For spatially varying albedo we use free textures from \cite{3dtextures}. Unlike \cite{xu2018deep,li2018learning,Boss2020-TwoShotShapeAndBrdf} we  do not use the roughness supplied with the textures because their details are too small to be recovered faithfully. Instead, we apply a random roughness to each primitive shape in the scene. For sampling roughness, we first sample a Phong exponent from an exponential distribution that is approximately the same as the one used by \cite{LIMEMeka:2018}.  We then convert it into a Beckmann equivalent roughness \cite{Mitsuba}. We observe that this design choice forces the network trained on this data to rely more on photometric cues than prior associations between albedo and roughness. Thus the network can generalize better to real data. More details on our synthetic data generation procedure can be found in Appendix \ref{data_generation}.

\subsection{Recursive Multi-Resolution Network}

Our goal is to predict high-resolution surface normals, albedo and roughness, such that when we integrate the normals to create a mesh, we can preserve the details of the object. An obvious choice is to train a convolutional network, e.g. ResNets \cite{CycleGAN2017}, on high-resolution input-output pairs. However, this is extremely difficult to achieve for the following reasons. First, the primitive shapes used for data generation lack details, and so do most synthetic shape datasets like Shapenet \cite{chang2015shapenet}. Thus the synthetic high-resolution data will lack the details often observed in real-world objects. Second, training a ResNet with sufficient depth on 1024$\times$1024 images will require significant memory and training time, which often makes it impossible to train on a single GPU with batch-size of one. ResNets trained on low-resolution 256$\times$256 images introduce additional artifacts when tested on 1024$\times$1024 images. Thus our goal is to design a network that can be trained at low-resolution yet still produce satisfactory results on high-resolution data during inference.

The input to our system is a set of images of an object illuminated by a light co-directional with the camera and optionally up to five additional images of the object illuminated by directional light from the right, center-right, center-left, left, and above. We also input a segmentation mask. Then our system predicts surface normal $N$, albedo $A$ and roughness $R$. We use an interactive segmentation tool \cite{fbrs2020_mask_making} to create a the mask, usually with just 3-5 clicks. 

Our network, depicted in Figure \ref{fig:network}, consists of two components, an initialization network (InitNet) and a recursive network (RecNet). InitNet receives input at a down-sampled resolution of 32$\times$32, $I^{32}$, and predicts normal, albedo, and roughness at the same resolution. This is used as initialization to the recursive network RecNet, which continuously upsamples the result by a factor of 2. RecNet starts by taking in the 32$\times$32 normal, albedo, and roughness predictions made by InitNet along with the input images downsampled to 64$\times$64 and predicts the normal, albedo, and roughness at 64$\times$64. These are then recursively fed into RecNet along with the inputs downsampled to 128$\times$128. This procedure is repeated until the normal, albedo, and roughness match the resolution of the original input images. This procedure is described in pseudo-code in Algorithm \ref{rec_algorithm}.

\begin{algorithm}
\caption{Network procedure: The input image at $2^{K} \times 2^{K}$ is down sampled to $2^k \times 2^k$ resolution, denoted by $I^{2^k}$, for $k$ from $K$ to $6$. $A^{2^k}$,$N^{2^k}$,$R^{2^k}$ represent albedo, normal and roughness predicted at $2^k \times 2^k$ resolution, respectively. }
\label{rec_algorithm}
\begin{algorithmic}
\STATE $N^{2^5},A^{2^5},R^{2^5} = InitNet(I^{2^5})$
\FOR {k = 6 to K} 
    \STATE $N^{2^k}, A^{2^k}, R^{2^k} = RecNet(I^{2^k},N^{2^{k-1}},A^{2^{k-1}},R^{2^{k-1}})$
\ENDFOR
\end{algorithmic}
\end{algorithm}

InitNet consists of three small Resnets, one for albedo, normal, and roughness, that only differ in the number of output channels: 3 for albedo; 2 for normal (we only predict the $x$ and $y$ component of the normal, $z$ is calculated as $\sqrt{1-x^2-y^2}$); and 1 for roughness. The ResNets in InitNet have no downsampling and only two residual blocks. The input to the network is a concatenation of six or fewer images and the mask of the object. When less than six images are used, respective channels in the input are zeroed. The RecNet architecture also consists of 3 ResNets; each ResNet contains 8 residual blocks, at its narrowest point each feature resolution is 1/4 the input resolution. The input to RecNet is six images and the mask along with the predicted normal, albedo and roughness from the previous step upsampled by a factor of 2. Please see Appendix \ref{network_architecture} for a detailed description of the network architectures.

Our architecture at test time applies the same network at each step to refine the network predictions from the previous step by a scale factor of 2. Thus we aim to train this network to be scale-independent. All networks are trained simultaneously. The networks are applied as in algorithm \ref{rec_algorithm}. This produces estimates of albedo, normal, and roughness at four resolutions from 32 to 256. We take the $L_1$ loss of normal, albedo, and roughness at all four resolutions and sum them to get our full loss function: 
\vspace{-0.5em}
\begin{equation*}
    \small{\min_{N,A,R} \sum_{k=5}^{8} L_1(A^{2^k},\hat{A}^{2^k}) +
    L_1(N^{2^k},\hat{N}^{2^k}) +
    L_1(R^{2^k},\hat{R}^{2^k})},
\vspace{-1em}
\end{equation*}
where $\hat{N}$, $\hat{A}$ and $\hat{R}$ denotes ground-truth normal, albedo, and roughness. Thus, RecNet is being trained at 3 scales simultaneously, and InitNet is only being trained at 32$\times$32.

\begin{figure}
    \centering
    \includegraphics[width=1.0\linewidth]{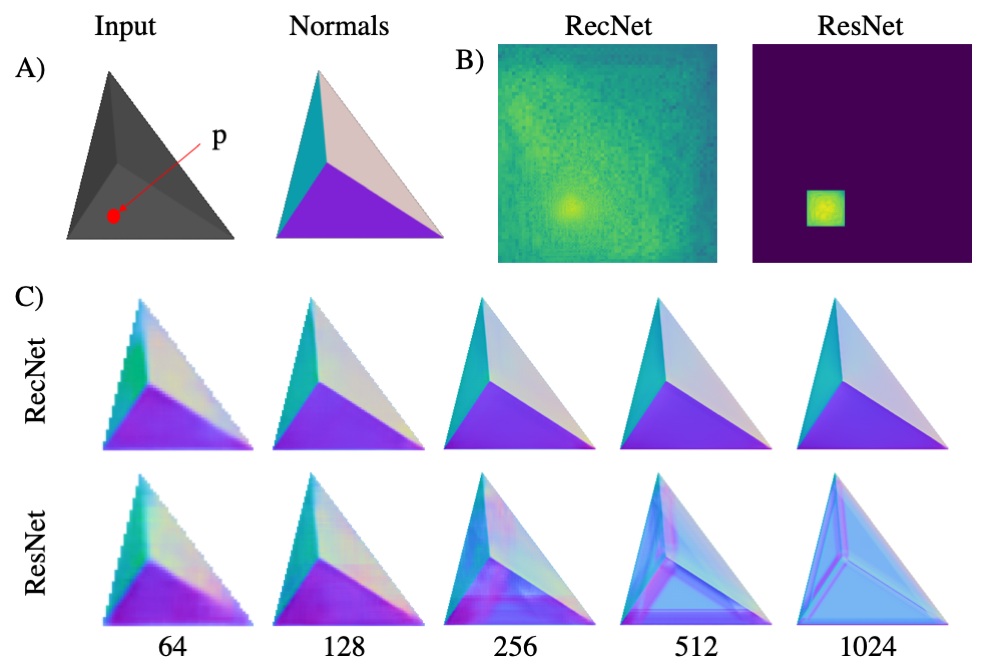}
    \caption{ (A) A pyramid rendered with co-directional light and camera, and its ground truth normals. (B) Visualization of the gradient of the input image with respect to output pixel p, for RecNet and ResNet at 512$\times$512 resolution. Both RecNet and ResNet are trained at 256$\times$256 and have receptive fields 93 and 85, respectively. Note that for the ResNet the output at pixel $p$ only depends on the input in a neighborhood of pixel $p$, this is not the case for RecNet. (C) Normal prediction by RecNet and ResNet at different input resolutions.}
    \label{fig:network_justification}
    \vspace{-1.5em}
\end{figure}

To explain the necessity of this recursive architecture, consider the rendering of a pyramid in Figure \ref{fig:network_justification}A. The pyramid is diffuse with uniform albedo, and the light is co-directional with the camera. If we would only consider a local window around point $p$ we would just see a uniform color; it requires global context to disambiguate the normal at this location. A fully convolutional architecture, such as a Resnet, has a finite receptive field; thus if we have this image at very high resolution, the network's receptive field will not be large enough to capture the global context necessary to disambiguate the normal. Furthermore, even if we trained a network with a very large receptive field, it seems unlikely that it could learn relationships more distant than the size of the training data itself, in our case 256$\times$256. 

In Figure \ref{fig:network_justification}C we see that as the resolution of the triangle grows, it requires more and more relatively distant information to predict the normal correctly; thus the quality of the prediction by the ResNet deteriorates. In contrast, with the RecNet the receptive field is proportional to the number of iterations i.e. image size; thus its prediction does not deteriorate with increasing image resolution. We show this in \ref{fig:network_justification}B by plotting the gradient of the input image with respect to pixel $p$. We observe that for the ResNet the normal at pixel $p$ is determined only by a small neighborhood around $p$ whereas the RecNet also takes into account pixels a much greater distance away. Although this example seems contrived, we show in section \ref{ablation} that for real data, the ResNet produces significantly more artifacts than the RecNet.

\subsection{Implementation}
We implement our network in Pytorch. We train with the Adam optimizer for 50 epochs with a batch size of 10 and a learning rate of $10^{-4}$. In each batch, one to five images out of six are randomly zeroed such that the probability of getting 1,2,3,4,5,6 non-zero images is 0.3, 0.2, 0.2, 0.1, 0.1, 0.1. Note the co-directional light image is never zeroed. Training took approximately two days on two Nvidia RTX2080Ti GPUs.

After our network predicts surface normal, albedo and roughness, we integrate the normals to obtain a depth map. Then we consider each pixel of the depth map as a vertex of a mesh. For details on our integration procedure, please see Appendix \ref{integrating_normals}.

%% file: compare_sota.tex
\begin{table*}[!h]
	\centering
	\caption{\textbf{Three Image Results on  DiLiGenT} Comparison of SDPS-Net \cite{chen2019SDPS_Net}, SDPS-Net retrained with 3 input images, and our method on DiLiGenT. Lights are from the front, front-right, and front-left of the camera. MAE (in degrees) for each object is reported. }
	\vspace{-0.5em}

		\begin{tabular}{|c|c|c|c|c|c|c|c|c|c|c|c|}
			\hline
			Algorithm & ball & cat & pot1 & bear & pot2 & buddha & goblet & reading & cow & harvest & mean                \\ \hline
			SDPS-Net \cite{chen2019SDPS_Net} & 15.4 & 22.4 & 25.7 & 17.9 & 18.5 & 24.0 & 31.0 & 29.3 & 21.4 & 27.1 & 23.3 \\ \hline
			SDPS-Net (3 image retrained) & \textbf{5.7} & 15.3 & 13.1 & 7.6 & 13.0 & 18.7 & 24.6 & 22.5 & 12.1 & 22.3 & 15.5 \\ \hline
        ResNet (ablation) & 5.8 & 14.7 & 12.9 & 8.3 & 12.3 & 14.9 & 23.4 & 17.3 & 15.7 & 23.2 & 14.8 \\ \hline
	    Ours & \textbf{5.7} & \textbf{12.7} & \textbf{10.6} & \textbf{7.1} & \textbf{10.2} & \textbf{13.9} & \textbf{17.0} & \textbf{16.9} & \textbf{9.9} & \textbf{20.6} & \textbf{12.5} \\ \hline
		\end{tabular}
	\label{tab:diligent_3image}
\end{table*}

\begin{table*}[!h]
	\centering
	\caption{\textbf{Single Image Results on  DiLiGenT} Comparison of
	\cite{chen2019SDPS_Net}, SDPS-Net retrained with 1 image (fully calibrated), \cite{li2018learning} and our method on DiLiGenT using a single input image with light approximately co-directional with the camera. MAE (in degrees) is reported.}
	\vspace{-0.5em}

		\begin{tabular}{|c|c|c|c|c|c|c|c|c|c|c|c|}
			\hline
			Algorithm & ball & cat & pot1 & bear & pot2 & buddha & goblet & reading & cow & harvest & mean                \\ \hline
			SDPS-Net \cite{chen2019SDPS_Net} & 36.0 & 35.4 & 36.3 & 34.2 & 36.7 & 44.3 & 43.2 & 43.4 & 35.4 & 42.0 & 38.7 \\ \hline
			SDPS-Net (1 image retrained) & 6.0 & \textbf{21.0} & 17.8 & 11.2 & 18.1 & 26.7 & 27.8 & 29.0 & 17.3 & 34.1 & 20.9 \\ \hline
			Li'18 \cite{li2018learning} & 20.4 & 29.7 & 19.5 & 27.2 & 20.2 & 32.1 & 22.6 & 32.3 & 21.4 & 37.1 & 26.3 \\ \hline
		    ResNet (ablation) & \textbf{5.4} & 22.8 & 16.4 & 9.2 & 15.1 & 23.6 & 27.8 & 24.6 & 15.5 & 30.5 & 19.1 \\ \hline
		    Ours & 7.1 & 21.1 & \textbf{12.7} & \textbf{8.3} & \textbf{12.7} & \textbf{20.7} & \textbf{20.3} & \textbf{22.3} & \textbf{11.7} & \textbf{29.9} & \textbf{16.7} \\ \hline
		\end{tabular}
         \vspace{-1.5em}
	\label{tab:diligent_1image}
\end{table*}

\section{Experiments}

We compare with various state-of-the-art approaches for shape and material estimation: $\bullet$ SPDS-Net \cite{chen2019SDPS_Net} is a PS technique that only predicts surface normals from multiple images. Although not emphasized in the original paper, it can also be used for predicting normals from a single image. SDPS-Net, like ours, can predict normals at 1024$\times$1024 resolution. $\bullet$ Li'18 \cite{li2018learning} predicts surface normal, albedo and roughness from a single image captured with co-directional camera and light at 256$\times$256 resolution. $\bullet$ Boss'20 \cite{Boss2020-TwoShotShapeAndBrdf} also predicts normal, albedo and roughness but requires two images, with and without flash, at 256$\times$256 resolution.

Since it is extremely difficult to capture ground-truth albedo and roughness of household objects, our evaluation relies on qualitative comparison. However, it is possible to quantitatively evaluate normals on real data with the DiLiGenT dataset \cite{diligent_data}. DiLiGenT consists of 10 objects with diverse materials captured under 96 calibrated lighting directions with ground truth normals. The lighting directions lie approximately on a rectangle and are all within roughly 45 degrees of the camera.

\begin{figure}
    \centering
    \includegraphics[width=1.0\linewidth]{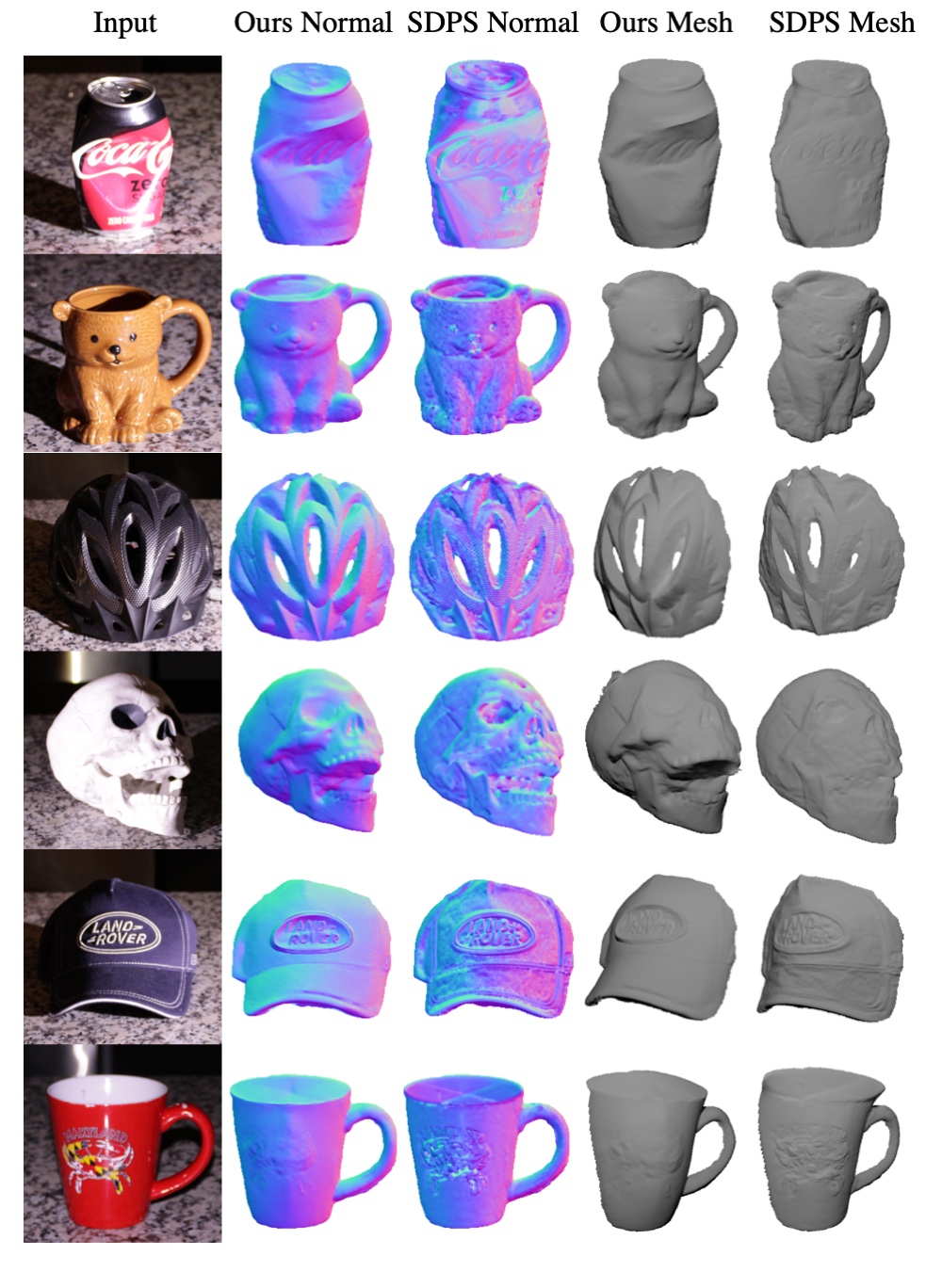}
    \vspace{-3em}
    \caption{Comparison of our normal estimation method with that of SDPS-Net retrained on 3 images. Tested with front, front-right and front-left images.}
    \label{fig:3 image comparison ours vs. SDPSnet retrained at 3}
    \vspace{-1em}
\end{figure}

\textbf{Surface Normal estimation with multiple images.} In Table \ref{tab:diligent_3image}, we compare our normal estimation with SDPS-Net on the  DiLiGenT dataset with three images coming from the front, front-right, and front-left. We are restricted to at most three images on this dataset because images from the left, right, and above are not included. We report a mean angular error (MAE) of 12.5$^{\circ}$ compared to 23.3$^{\circ}$ for \cite{chen2019SDPS_Net}. However, SDPS-Net is trained on 32 images,  so we retrain it with only three images as input. Although this improves the quality of SDPS-Net our method is still numerically superior to it. Furthermore, the qualitative evaluation in Figure \ref{fig:3 image comparison ours vs. SDPSnet retrained at 3}, clearly shows the superior performance of our method on challenging objects. We also show superior performance to SDPS-Net on the two image task (see Appendix \ref{compare_sdpsnet}).

\begin{figure}
    \centering
    \includegraphics[width=1.0\linewidth]{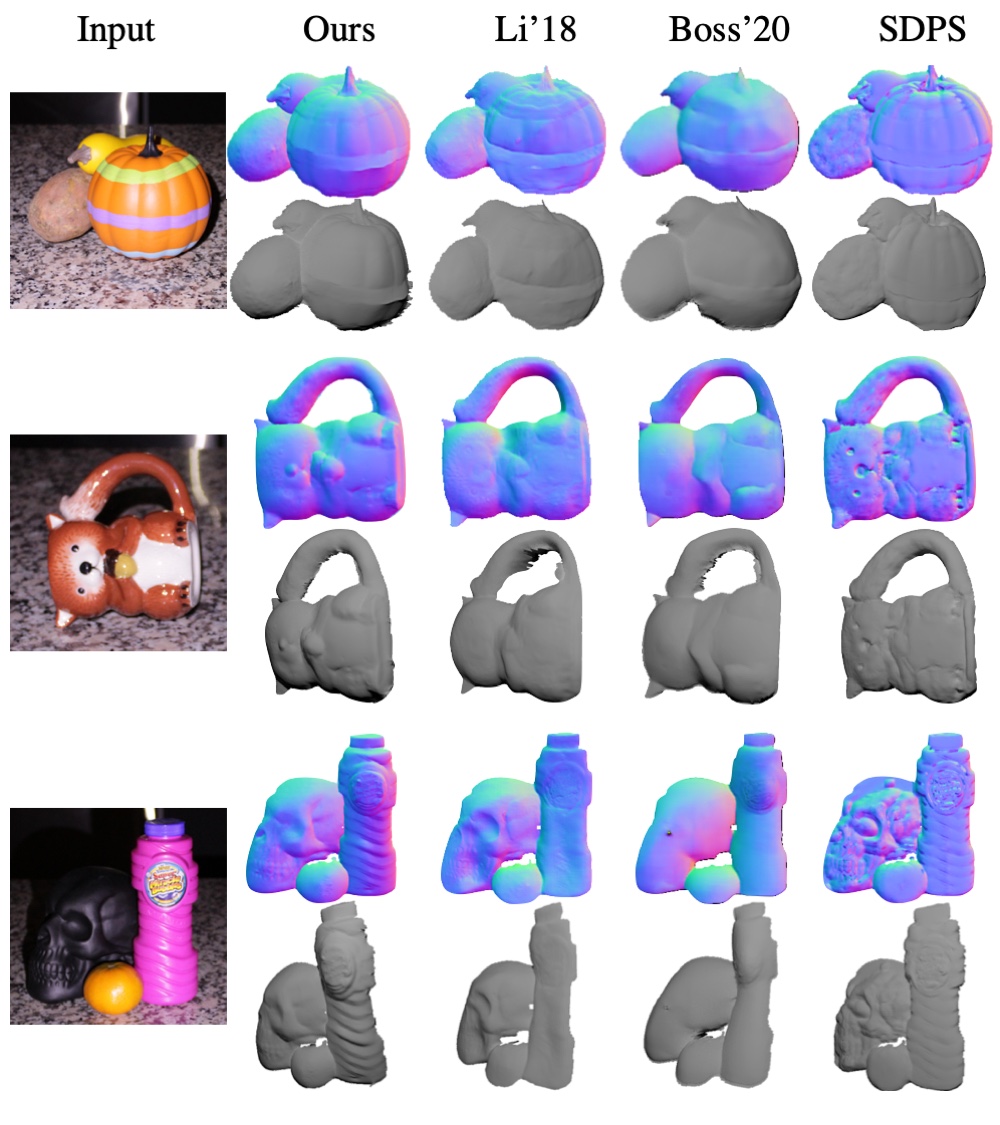}
    \vspace{-3em}
    \caption{Comparison of normal estimation with Li'18 \cite{li2018learning}, Boss'20 \cite{Boss2020-TwoShotShapeAndBrdf} and SDPS-Net retrained on a single image \cite{chen2019SDPS_Net}.}
    \vspace{-2em}
    \label{fig:1_image_compare_normal}
\end{figure}

\textbf{Surface Normal estimation with single image.} In Table \ref{tab:diligent_1image} we compare normal estimation error of our method with that of Li'18 \cite{li2018learning} and SDPS-Net \cite{chen2019SDPS_Net} on  DiLiGenT using a single image captured with approximately co-directional light and camera. We also retrained SDPS-Net just on a single image, although our method is trained to work with up to 6 images at once, and still, we outperform SDPS-Net. We could not compare with Boss'20 \cite{Boss2020-TwoShotShapeAndBrdf} on DiLiGenT as it requires two images with and without flash. In Figure \ref{fig:1_image_compare_normal}, we present a qualitative comparison of our method with that of SDPS-Net retrained on a single image, Li'18 and Boss'20 on images captured by us. Both quantitative and qualitative evaluations show that our method outperforms state-of-the-art normal estimation techniques for single and few image inputs.

\begin{figure}
    \centering

    \includegraphics[width=1.0\linewidth]{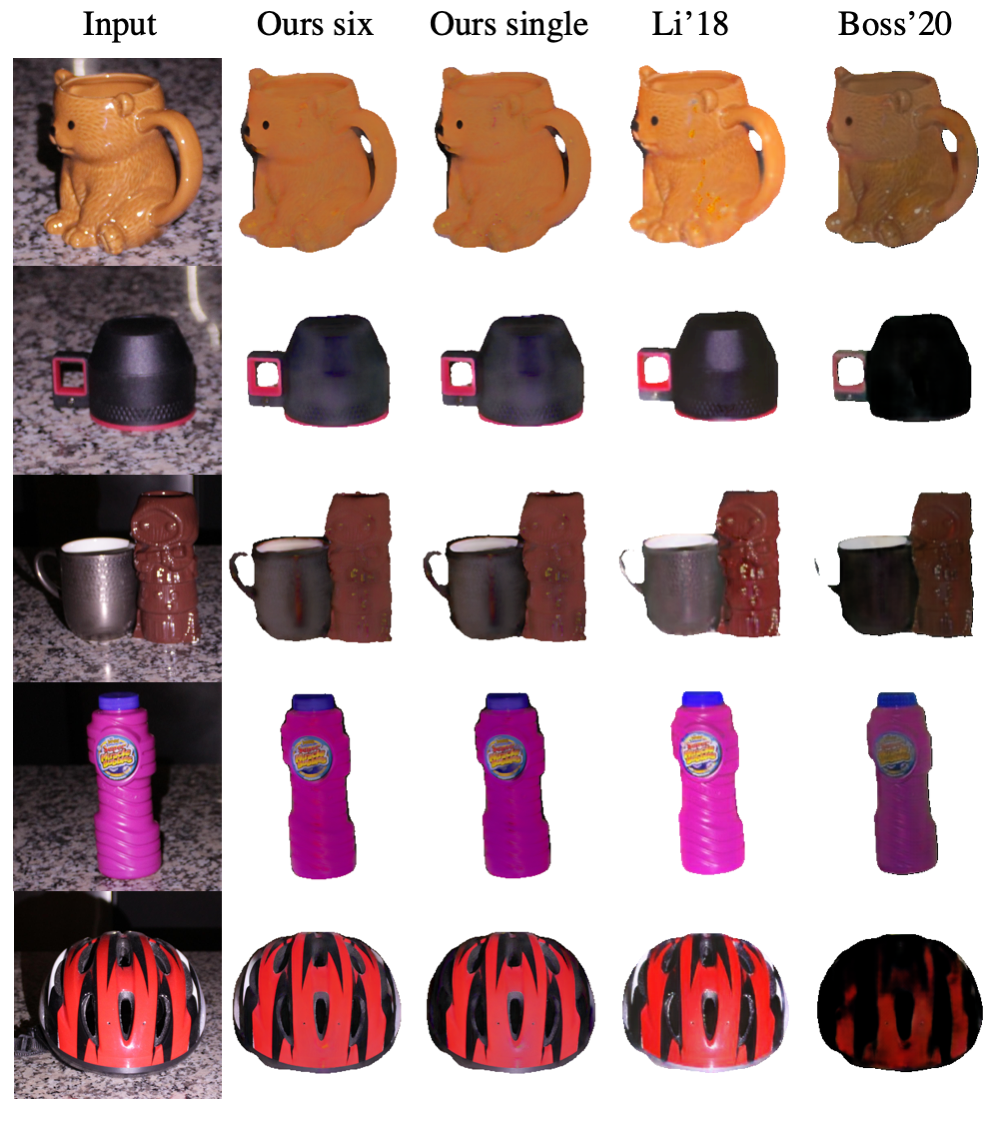}
    \vspace{-2.5em}
    \caption{Comparison of albedo estimation between Li'18 \cite{li2018learning}, Boss'20 \cite{Boss2020-TwoShotShapeAndBrdf}, our method with one image (ours single) and our method with six images (ours six).}
    \vspace{-2em}
    \label{fig:1_image_compare_albedo}
\end{figure}

\begin{figure}
    \centering
    \includegraphics[width=1.0\linewidth]{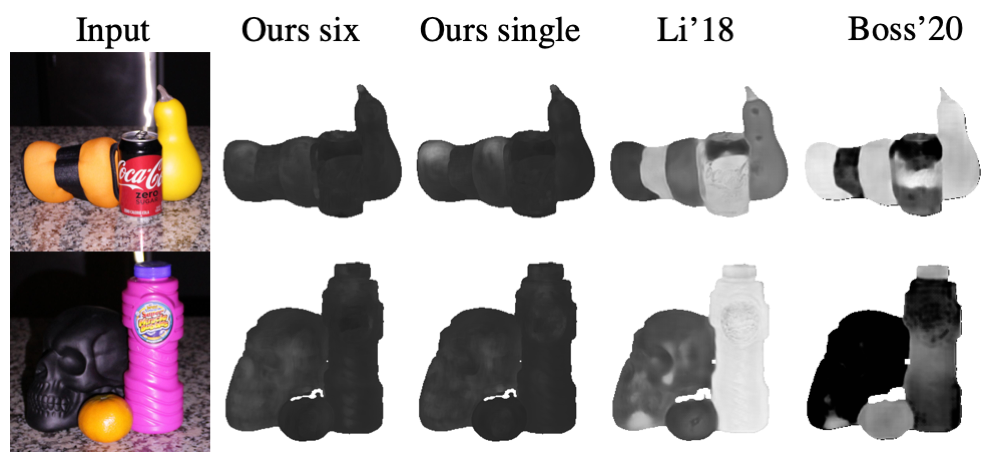}
    \vspace{-2em}
    \caption{Comparison of roughness estimation between Li'18 \cite{li2018learning}, Boss'20 \cite{Boss2020-TwoShotShapeAndBrdf}, our method with one image and our method with six images. Brighter indicates rougher (less specular).}
    \label{fig:roughness}
    \vspace{-1em}
\end{figure}

\textbf{Material Estimation.} We present a visual comparison of albedo in Figure \ref{fig:1_image_compare_albedo} and roughness in Figure \ref{fig:roughness} estimated by our method with that of Li'18 using a single image and Boss'20 with the same image and an additional image without directional light. We also show the albedo and roughness predicted by our method with six images as input. We observe that our method is often better in removing cast shadows and specular highlights from the albedo. Roughness is challenging for all methods. We often observe that even relative roughness predictions do not agree between methods. More qualitative results on our data and data collected by \cite{li2018learning} can be found in Appendix \ref{ravi_results}.

%% file: ablation.tex
\section{Ablation Studies}
\label{ablation}

\begin{figure}
    \centering
    \includegraphics[width=1.0\linewidth]{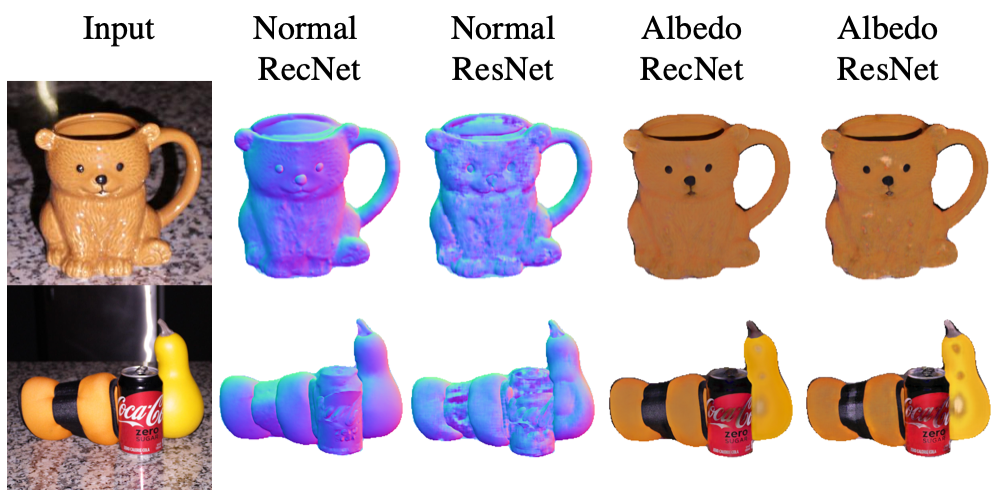}
    \vspace{-1.5em}
    \caption{Illustration of artifacts produced by ResNet when trained at 256$\times$256 and tested at 1024$\times$1024 compared to RecNet.}
    \label{fig:resnet_artifacts}
    \vspace{-0.5em}
\end{figure}

\textbf{Evaluation of Recursive Architecture.} To evaluate the effectiveness of our recursive architecture, we train a ResNet based on \cite{CycleGAN2017,isola2017image}. Our recursive network has approximately 14.5M parameters, and the vanilla Resnet has approximately 16M parameters. Both our recursive network, RecNet, and ResNet are trained at 256$\times$256 resolution and tested on 1024$\times$1024. Tables \ref{tab:diligent_3image} and \ref{tab:diligent_1image} show quantitatively that our recursive architecture outperforms the ResNet on DiLiGenT in the 3 image and 1 image cases, respectively. Figure \ref{fig:resnet_artifacts} demonstrates ResNet introducing artifacts when tested at 1024$\times$1024.

\textbf{Training with a Fixed Number of Inputs.} We also trained specific variants of our network that only use a fixed number of images as input, either one or three, rather than a random number of inputs during training. This means that instead of having one network that can handle any number of inputs, we obtain separate network weights for different numbers of input images. We found that our proposed training method is only slightly worse ($< 2^{\circ}$) on DiLiGenT than networks trained separately for 1 and 3 image inputs. This is shown in Table \ref{tab:ablation_fixed_num_inputs}. However, we felt that having a single network that performs well across different numbers of input images is more desirable than having different networks for different numbers of input images.

\begin{table}[t]
  \begin{center}
    \setlength\tabcolsep{2 pt}
    \begin{tabular}{l|rr|rr|}
      \toprule
      \multicolumn{1}{c|}{} & \multicolumn{2}{c|}{3 image input} & \multicolumn{2}{c|}{1 image input} \\
       & Ours & Ours-3 & Ours & Ours-1 \\
      \midrule
      MAE (in degrees) & 12.5 & 10.7 & 16.7 & 16.1 \\
      \bottomrule
    \end{tabular}
    \vspace{-0.5em}
    \caption{Our network can handle any number of input images (1-6), and performs comparable to training separate networks with a fixed number of input images; Ours-3 for 3 image input network and Ours-1 for single image input network.}
    \label{tab:ablation_fixed_num_inputs}
  \end{center}
      \vspace{-2.5em}
\end{table}

\textbf{Robustness to Angle and Intensity Variation.} Although we need lighting to be weakly calibrated, our method is quite robust to variation in capture directions. We test this by computing the MAE on DiLiGenT, as the input lighting directions move further away from our prescribed directions. Since our front-right and front-left directions are at the extremes of the range present in DiLiGenT, we start from these extremes and move progressively inward. We leave the center image fixed for this experiment. Table \ref{tab:angular_deviation} shows a minimal increase in error while the lights are within about $12.5^\circ$ of the optimal.

We also demonstrate our method's robustness to intensity variation using DiLiGenT. To simulate variation in intensity, we multiply each HDR image in DiLiGenT by a random scalar sampled from a Gaussian with unit mean and various standard deviations before tonemapping and clamping between 0 and 1. We repeat the experiment five times at each standard deviation. Table \ref{tab:intensity_deviation} shows the error at each standard deviation averaged over the five runs. We see that only at a high standard deviation, above about 0.8, when many pixel values are equal to 0 and 1, do we get a significant increase in normal prediction error.

\begin{table}[]
    \centering
    \begin{tabular}{|c|c|c|c|c|c|}
     \hline
        Deviation (degrees) &  4.8 & 8.2 & 12.2 & 16.7 & 21.6 \\ \hline 
       MAE & 12.5 & 12.4 & 12.7 & 13.4 & 15.0
        \\ \hline
    \end{tabular}
    \vspace{-0.5em}
    \caption{Error (MAE in degrees) on the DiLiGenT dataset as the direction of the light deviates from the optimal.}
    \label{tab:angular_deviation}
    \vspace{-0.5em}
\end{table}

\begin{table}[]
    \centering
    \begin{tabularx}{\columnwidth}{|c|c|c|c|c|c|c|}
    \hline
        s.d. scaling & 0 & 0.1 & 0.2 & 0.4 & 0.8 & 1.2 \\ \hline
       MAE & 12.5 & 12.5 & 12.5 & 12.9 & 14.6 & 16.2 
   \\ \hline
    \end{tabularx}
        \vspace{-0.5em}
    \caption{Error (MAE in degrees) on the DiLiGenT dataset as the intensity changes based on standard deviation (s.d.) of a unit mean gaussian.}
    \vspace{-1.0em}
    \label{tab:intensity_deviation}
\end{table}

\begin{figure}
    \centering
    \includegraphics[width=1.0\linewidth]{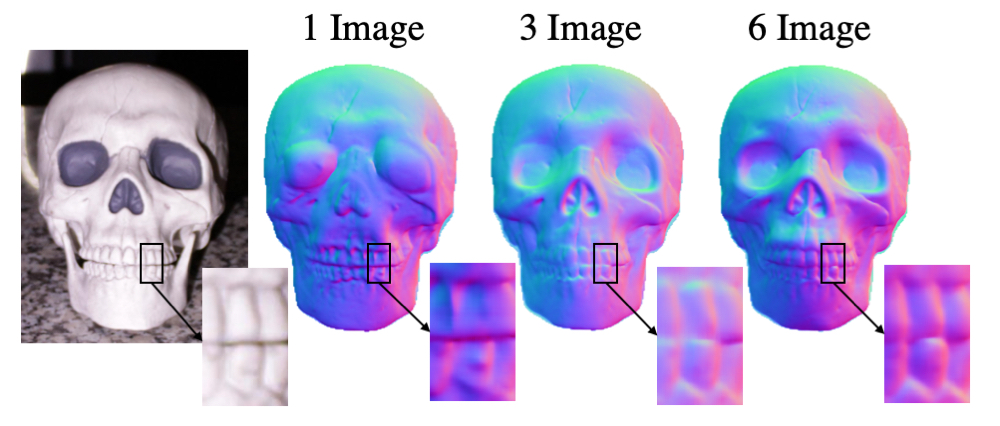}
    \vspace{-2em}
    \caption{Results of our network with 1, 3 and 6 input images.}
    \label{fig:progression}
    \vspace{-1.5em}
\end{figure}

\textbf{Effect of Number of Input Images.} 
Figure \ref{fig:progression} shows how normal quality improves with more input images.

%% file: conclusion.tex
\vspace{-0.5em}
\section{Conclusion}
Shape and reflectance capture is a fundamental research problem in Computer Vision and Graphics, with applications in AR/VR, e-commerce etc. Our goal is to create a technique that allows users to capture high-quality shape and reflectance of an object in a household setting, with a camera, a flashlight, and ideally a tripod. Our technique relies on weakly calibrated capture done with a handheld flashlight. Our method is robust up to 10-15$^\circ$ error in lighting direction. Weakly calibrated photometric stereo, even with error in lighting direction, appears to produce better reconstruction than uncalibrated Photometric Stereo.

Our main technical novelty is the recursive neural network RecNet, which can predict geometry and reflectance at $2^{k}\times2^{k}$ resolution given an input image at $2^{k}\times2^{k}$ and estimated geometry and reflectance from the previous step at $2^{k-1}\times2^{k-1}$. This allows us to train at 256$\times$256 and test at higher resolution, such as 1024$\times$1024. Training a neural network directly on higher resolution data is extremely difficult due to computational bottlenecks such as memory and training time and lack of large scale high-resolution synthetic data. Previous architectures suffer poor generalization when they train at low-resolution and perform inference at higher resolution due to a limited receptive field that does not capture global context. Our recursive architecture has a unique property in which the receptive field is doubled at each step, thus effectively allowing it to capture global information even with high-resolution images.

\textbf{Acknowledgment} This research is supported by the NSF under grant no. IIS-1526234 and IIS-1910132. 

\clearpage
\newpage

%% file: appendix.tex
\section{Appendix}
\label{appendix}
In this appendix, we provide additional details and results on our method.

\subsection{BRDF Model}
\label{brdf_model}
We define SV-BRDF using the Cook-Torrance model \cite{cook-torrance}, where the BRDF $B(V,L)$ is a 4D function of viewing direction $V$ and lighting direction $L$. $N$ is the surface normal, $H$ is the halfway vector of $V$ and $L$,  $\theta_X$ is $\arccos(X \cdot N)$, $A$ is the albedo and $R$ is the roughness. The model is then defined by the equations:  

\begin{equation}
    B(V,L) = \frac{A}{\pi} + \frac{F(H,V)D(H,N)G(H,L,V)}{4 (N \cdot L) (N \cdot V)}.
    \label{eq:fres}
\end{equation}

\begin{equation}
    D(H,N) = \frac{1}{\pi R^2 \cos^4(H \cdot N)} e^{\frac{-\tan^2(H \cdot N)}{\alpha^2}}.
\end{equation}

\begin{equation}
F(H,V) = F_0 + (1-F_0)(1-(H \cdot V))^5
\end{equation}

\begin{equation}
    a(X) = \frac{1}{R \tan{\theta_X}}
\end{equation}

\begin{equation}
    G1(X,H) =
    \begin{cases}
      1 & \hspace{-20mm} \text{if} a > 1.6 \\
      0  & \hspace{-20mm} \text{if} (H \cdot X)(X \cdot N) \leq 0 \\
      \frac{3.535a(X) + 2.181\sqrt{a(X)}}{1+2.276a(X) + 2.577\sqrt{a(X)}} & \text{otherwise}
    \end{cases}
\end{equation}

\begin{equation}
    G(H,L,V) = G1(L,H)G1(V,H)
\end{equation}

$F(H,V)$ is the Schlick approximation to the Fresnel term, $G(H,L,V)$ is Smith's masking-shadowing function with fast rational function approximation \cite{Mitsuba2}, and $D(H,N)$ is the Beckmann distribution. We fix the Fresnel-Schlick $F_0$ value at 0.05 as done by \cite{li2018learning}. Thus BRDF can be parametrized with albedo $A$ and roughness map $R$.

\subsection{Synthetic Data Generation and Augmentation}
\label{data_generation}

\subsubsection{Random BRDF Generation}
For spatially varying albedo we gathered 415 free textures from \cite{3dtextures}, which we divided into train and test sets with a 90-10 split. We augment these albedos, at render time, by multiplying each channel by a random Gaussian variable with mean 1 and standard deviation 0.2. 

For generating roughnesses, we take a similar approach to \cite{LIMEMeka:2018}. In \cite{LIMEMeka:2018} they randomly sample Phong exponents uniformly from the bins 0–10, 10–20, 20–40, 40–80, 80–160, 160–320, 320–640,640–2560. We approximate this by sampling from an exponential distribution with median 80. We then convert the sampled Phong exponent to a Beckmann equivalent roughness, $R$, with the formula $R= \sqrt{\frac{2}{2+E}}$, where $E$ is the Phong exponent as suggested by \cite{Mitsuba}.

We found that the Fresnel term does not make a large difference in appearance visually, so to simplify things we fix $F_0 = 0.05$ as done by \cite{li2018learning}.

\subsubsection{Scene Generation}
We create two synthetic datasets to train our model. The data generation procedure is the same for both datasets, the difference being that the first dataset uses random geometry generated by \cite{xu2018deep} and the second uses realistic geometry from the sculptures dataset \cite{sculpture_data}. The first dataset has 20,000 training scenes and 500 test scenes. The second has 10,000 training scenes and 200 test scenes.

The scene layout consists of a rectangle in the x-z plane to represent a floor and either a shape randomly selected from the 5000 shapes generated by \cite{xu2018deep}, which are composed of 1-9 primitives, or a shape from the sculpture dataset. Each primitive, including the floor, is assigned a random BRDF using the procedure described above. An orthographic camera is placed in the y-z plane, pointing at the shape and making an angle with the floor that is randomly sampled between 10$^\circ$ and 45$^\circ$.

The scene is rendered with six directional lights with unit intensity. The right, front-right, front-left and left lights are first placed with azimuth angles -90$^\circ$,-45$^\circ$,45$^\circ$,90$^\circ$ and elevations of 25$^\circ$ above the floor. They are then perturbed in both the azimuth and elevation randomly by up to 10$^\circ$. The overhead light is placed with random elevation between 80$^\circ$ and 90$^\circ$ above the floor and random azimuth between 0$^\circ$ and 360$^\circ$. The co-directional light is simply placed along the camera optical axis.

Scenes are rendered with Mitsuba2's \cite{Mitsuba2} Path-Tracer at 256 samples per pixel in HDR. Our BRDF implementation is a Mitsuba2 port of Boss's Mitsuba1 code \cite{Boss2020-TwoShotShapeAndBrdf}.

\subsubsection{Training Time Augmentation}
At training time images undergo one of two possible size transforms. With probability 0.7 they are randomly cropped to between 70\% and 100\% of their initial size and resized back to 256$\times$256, and with probability 0.3 they are randomly resized to between 60\% and 100\% of their initial size. They are then padded with zeros back to a size of 256$\times$256. We performed this procedure so that the network will see the same features at various scales.

We simulate intensity variations by randomly scaling each linear image to have a median selected randomly between 0.01 and 0.2. The images are then sRGB tonemapped and clamped between 0 and 1 before being fed to the network. We found this gives the images intensity variations fairly similar to those observed in natural images.

\subsection{Network Architectures}
\label{network_architecture}

\subsubsection{Notation}
To describe the network architectures used in this paper we first define some notation:
\begin{itemize}
    \item A-B $\coloneqq$ apply layer A then apply layer B
    \item BN $\coloneqq$ batch norm
    \item Relu $\coloneqq$ relu activation
    \item conv\_kn\_fm\_sp $\coloneqq$ convolution layer with kernel of size n and m filters (i.e. output channels) and stride p. If stride is 1 we will omit \_s1.
    \item convt\_kn\_fm\_sp $\coloneqq$ transposed convolution layer with kernel of size n and m filters (i.e. output channels) and stride p.
    \item Res\_n $\coloneqq$ conv\_k3\_fn - BN - Relu - conv\_k3\_fn - BN - +input. This defines the residual block.  +input indicates adding the input value to the output value.
\end{itemize}

\subsubsection{InitNet}
InitNet consists of 3 separate networks: one for albedo, normal, and roughness. We will call these InitNetModules. Each InitNetModule takes in 19 channels that are the concatenation of the six three channel images and the segmentation mask. The InitNetModule architecture is given by:
\begin{itemize}
    \item InitNetModule\_c $\coloneqq$ conv\_k7\_f64 - BN - Relu - Res\_64 - Res\_64 - conv\_k7\_fc.
\end{itemize}

Where c is 3, 2 and 1 for albedo, normal, and roughness, respectively.

\subsubsection{RecNet}
Similarly to InitNet, RecNet consists of three RecNetModules. Each takes in 25 channels: 19 for the images and masks, and 6 for the albedo, normal, and roughness estimated by the previous step, which are upsampled by a factor of two to match the size of the input images. The RecNetModule structure is defined as:

\begin{itemize}
    \item RecNetModule\_c $\coloneqq$ conv\_k7\_f64 - BN - Relu - Res\_64 - Res\_64 - conv\_k3\_f128\_s2 - BN - Relu - Res\_128 - Res\_128 - conv\_k3\_f256\_s2 - Res\_256 - Res\_256 - convt\_k3\_f128\_s2 - Res\_128 - Res\_128 - convt\_k3\_f64\_s2 - BN - Relu - conv\_k7\_fc
\end{itemize}

Where c is 3, 2 and 1 for albedo, normal, and roughness, respectively. A diagram of the InitNet-RecNet application is included in Figure \ref{fig:network} for reference.

\subsubsection{ ResNet for Ablation}
For the ablation study we used what is referred to as ResNet in the paper. This is actually 3 separate ResNets, one for albedo, normal, and roughness; although they are trained simultaneously. Their architectures are all the same and are based on  \cite{CycleGAN2017}. For consistence we refer to these ResNets as ResNetModules. Their architecture is given by:
\begin{itemize}
    \item ResNetModule $\coloneqq$ conv\_k7\_f64 - BN - Relu - conv\_k3\_f128\_s2 - BN - Relu - conv\_k3\_f256\_s2 - BN - Relu - Res\_256 - Res\_256 - Res\_256 - Res\_256 - convt\_k3\_f128\_s2 - BN - Relu - convt\_k3\_f64\_s2 - BN - Relu - conv\_k7\_fc
\end{itemize}

Where c is 3, 2 and 1 for albedo, normal, and roughness, respectively.

\subsection{SDPS-Net Retraining}
\label{sdps_net_retraining}
SDPS-Net \cite{chen2019SDPS_Net} consists of two networks: LCNet which takes $n$ images of an object under varying lighting and estimates the lighting conditions for each image, and NENet which takes the same $n$ images as well as the estimated lighting directions and predicts the normals. Although in principle these networks can take an arbitrary number of input images, we found that performance decreases significantly if the number of training images differs greatly from the number of test images. Therefore, to give SDPS-Net a fair chance, we retrained SDPS-Net three times with 1,2 and 3 input images at training time. For the case of 2 and 3 input images this consisted of a full retraining of LCNet and NENet using the author's default parameters. In the one image case, we only trained NENet by providing it with the ground truth lighting directions rather than those predicted by LCNet. 

\subsection{Integrating Normal Maps}
\label{integrating_normals}
Let $f(x,y)$ be the depth of the surface at pixel location $(x,y)$, then the surface normal is given by $(n_1,n_2,n_3) = \frac{1}{\sqrt(f_x^2 + f_y^2 + 1)}(f_x,f_y,-1)$. So to find $f$ we can solve the system.

 \begin{equation}
     f_x = \frac{-n_1}{n_3}
 \end{equation} 
 \begin{equation}
     f_y = \frac{-n_2}{n_3}
 \end{equation}
 We discretize this as:
 
 \begin{equation}
     f(x_{i+1},y_j) - f(x_i,y_j) = \frac{-n_1(x_i,y_j)}{n_3(x_i,y_j)}
 \end{equation}
 \begin{equation}
     f(x_i,y_{j+1})-f(x_i,y_j) = \frac{-n_2(x_i,y_j)}{n_3(x_i,y_j)}
 \end{equation}
 
 Where $(i,j)$ are the pixel indices. For a normal map with $n$ pixels these equations describe a sparse linear system of $2n$ equations in $n$ unknowns. We find a least squares solution to this system using using Pytorch's LBGFS optimizer.

\subsection{Comparison to SDPSNet (2 Image)} 
\label{compare_sdpsnet}
In Tables \ref{tab:diligent_2image_left} and \ref{tab:diligent_2image_right} we compare our method to SDPS-Net \cite{chen2019SDPS_Net}, and SDPS-Net retrained with 2 input images on DiLiGenT in the case where the network is given two input images. For the inputs we use either the images from the front and front-left \ref{tab:diligent_2image_left} or front and front-right \ref{tab:diligent_2image_right}.  We show superior performance to SDPS-Net even when it is retrained specifically for two input images.

\begin{table*}[ht]
	\centering
	\caption{\textbf{Two Image Results on  DiLiGenT (Left)} Comparison of SDPS-Net \cite{chen2019SDPS_Net}, SDPS-Net retrained with 2 input images, and our method on DiLiGenT, using images front and front-left. MAE (in degrees) for each object is reported. }
	\vspace{-0.5em}

		\begin{tabular}{|c|c|c|c|c|c|c|c|c|c|c|c|}
			\hline
			Algorithm & ball & cat & pot1 & bear & pot2 & buddha & goblet & reading & cow & harvest & mean                \\ \hline
			SDPS-Net & 25.3 & 27.4 & 29.5 & 23.7 & 24.0 & 31.7 & 36.7 & 35.0 & 28.9 & 31.7 & 29.4 \\ \hline
			SDPS-Net (retrained)  & 5.7 & 17.1 & 14.9 & 8.7 & 16.3 & 20.3 & 25.2 & 24.6 & 12.7 & 26.1 & 17.2
		 \\ \hline
            Ours & 6.2 & 14.7 & 11.4 & 7.7 & 11.2 & 15.0 & 18.7 & 17.7 & 9.7 & 24.7 & 13.7 \\ \hline
		\end{tabular}
	\label{tab:diligent_2image_left}
\end{table*}

\begin{table*}[ht]
	\centering
	\caption{\textbf{Two Image Results on  DiLiGenT (Right)} Comparison of SDPS-Net \cite{chen2019SDPS_Net}, SDPS-Net retrained with 2 input images, and our method on DiLiGenT, using images front  and front-right. MAE (in degrees) for each object is reported. }
	\vspace{-0.5em}

		\begin{tabular}{|c|c|c|c|c|c|c|c|c|c|c|c|}
			\hline
			Algorithm & ball & cat & pot1 & bear & pot2 & buddha & goblet & reading & cow & harvest & mean                \\ \hline
			SDPS-Net & 27.0 & 31.0 & 33.8 & 24.2 & 26.0 & 30.8 & 41.8 & 37.1 & 29.5 & 31.6 & 31.3 \\ \hline
			SDPS-Net (retrained) & 6.3 & 19.0 & 17.4 & 9.1 & 15.8 & 21.1 & 27.5 & 24.6 & 14.4 & 26.5 & 18.2 \\ \hline
	        Ours & 6.8 & 14.9 & 11.7 & 7.7 & 10.9 & 15.3 & 19.1 & 19.7 & 10.7 & 24.0 & 14.1 \\ \hline
		\end{tabular}
	\label{tab:diligent_2image_right}
\end{table*}

\subsection{Results on Data Captured by Li~\cite{li2018learning}}
\label{ravi_results}
\begin{figure*}

\includegraphics[width=0.9\textwidth]{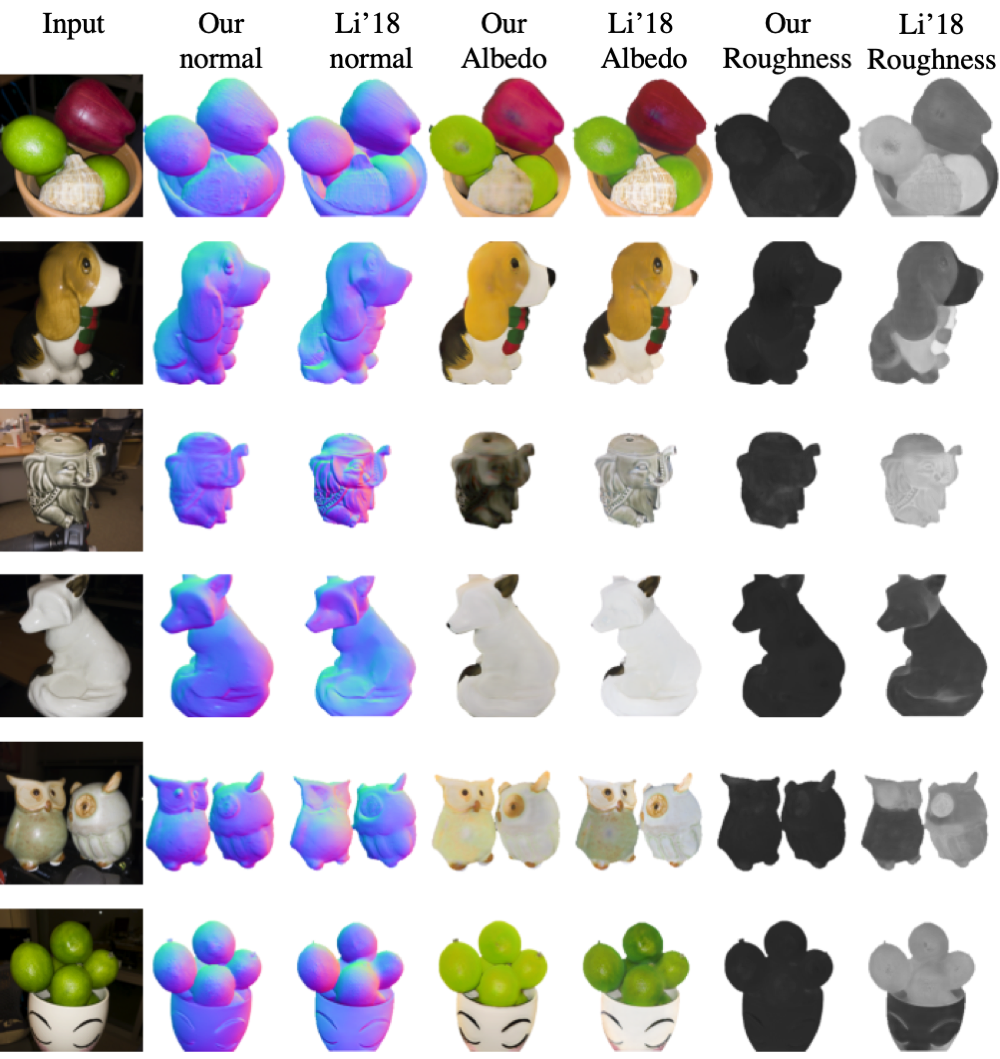}

\caption{Comparison of our single image results vs. those of Li'18 \cite{li2018learning} on data captured by Li'18.}
\label{ravi_data_a}
\end{figure*}

\begin{figure*}
\includegraphics[width=0.9\textwidth]{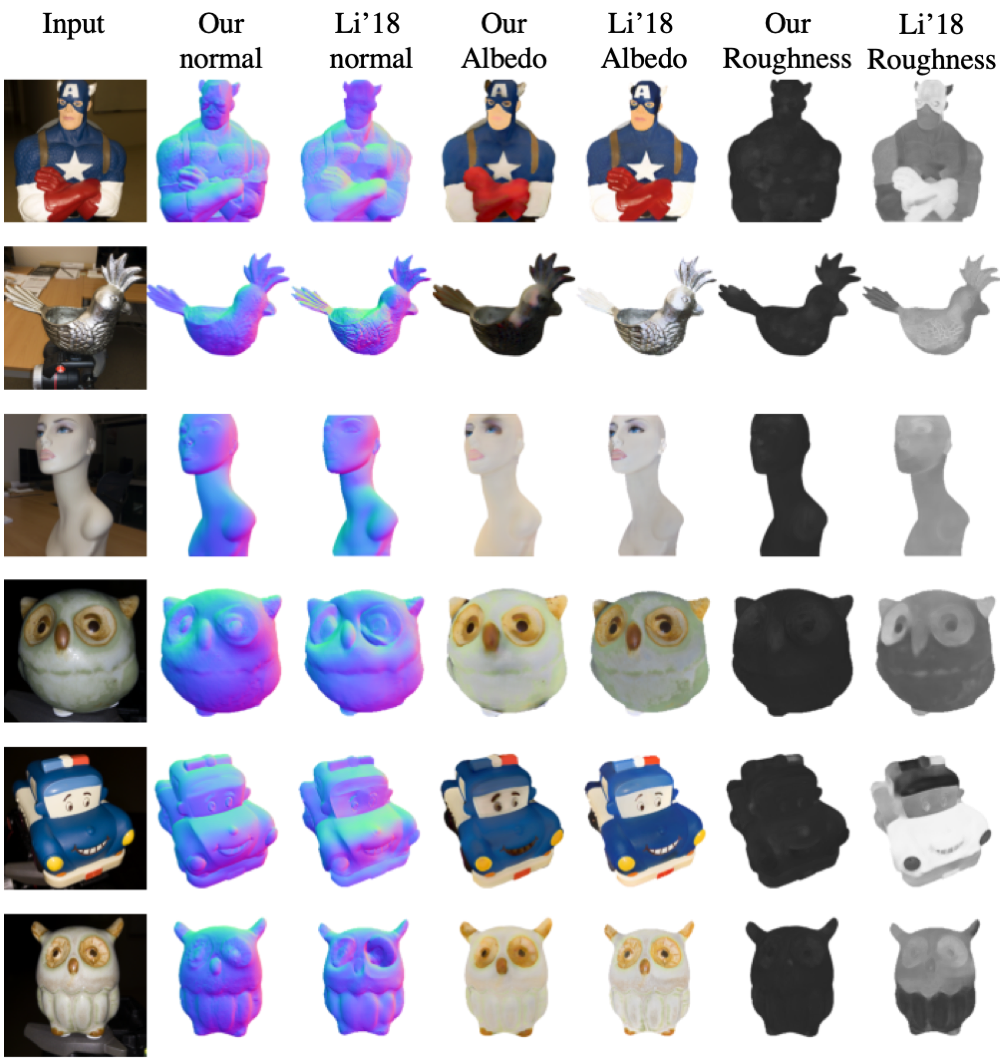}

\caption{Comparison of our single image results vs. those of Li'18 \cite{li2018learning} on data captured by Li'18.}
\label{ravi_data_b}
\end{figure*}

In Figures \ref{ravi_data_a} and \ref{ravi_data_b} we compare our results for albedo, normal, and roughness estimation to those of \cite{li2018learning} on the data captured by \cite{li2018learning}.

\subsection{More Results on Our Data}
\label{our_results}

\begin{figure*}
    \centering
    \includegraphics[width=1\textwidth]{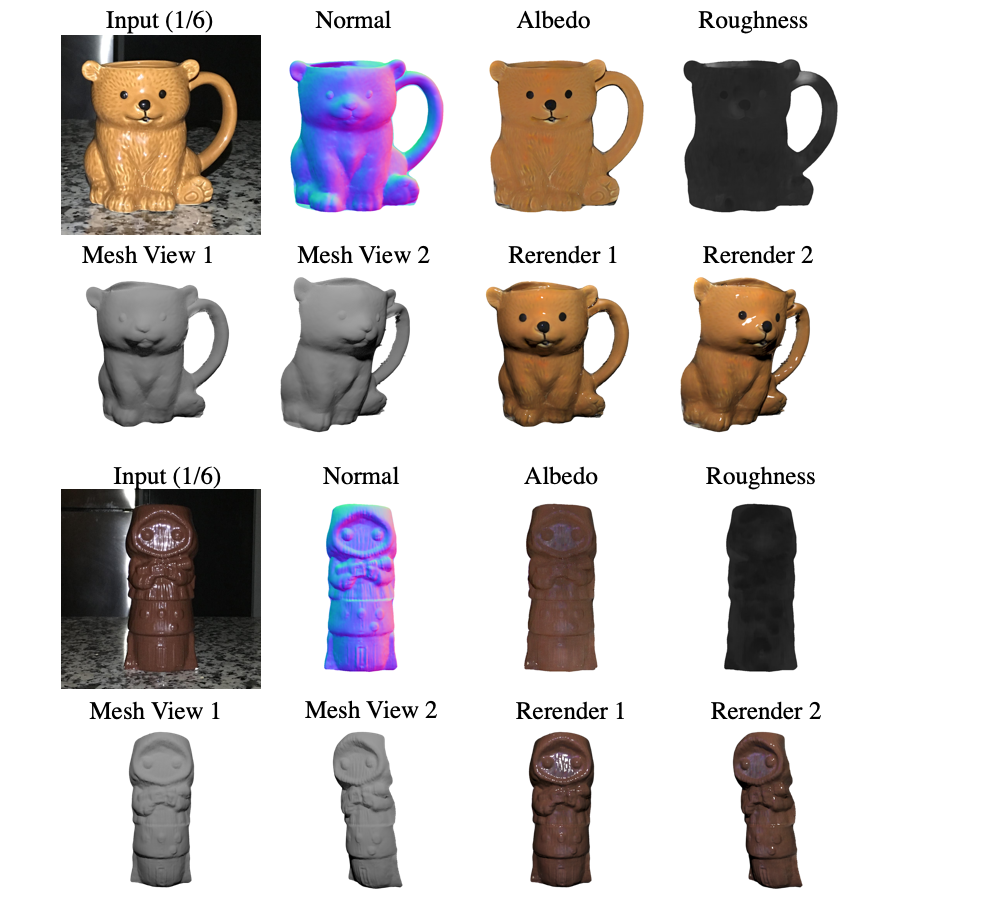}
    \caption{Results taken with only an iPhone, flashlight and improvised stand. }
    \label{fig:minimal_setup}
\end{figure*}

In this section, we show more comparisons to state-of-the-art methods for multi-image normal estimation and single image normal, albedo, and roughness estimation on the data we captured. We also show more results on images captured with our minimal setup using only an iPhone, flashlight, and improvised stand in Figure \ref{fig:minimal_setup}.

\begin{figure*}
    \centering
    \includegraphics[width=1\textwidth]{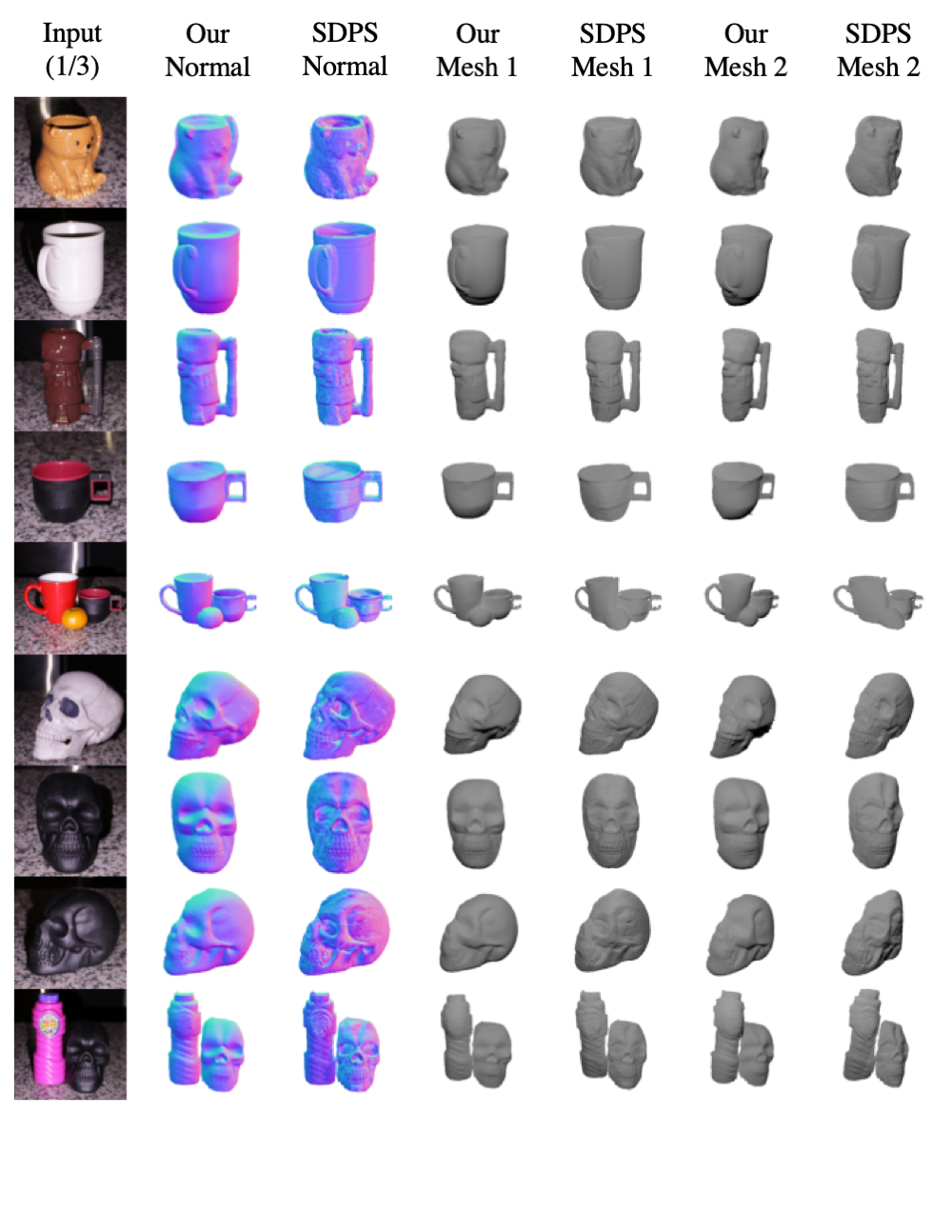}
    \vspace{-8em}
    \caption{Normal comparison with 3 input images between our method and SDPS-Net \cite{chen2019SDPS_Net} retrained with 3 input images.}
    \label{fig:multi_image_normal_a}
\end{figure*}

\begin{figure*}
    \centering
    \includegraphics[width=1\textwidth]{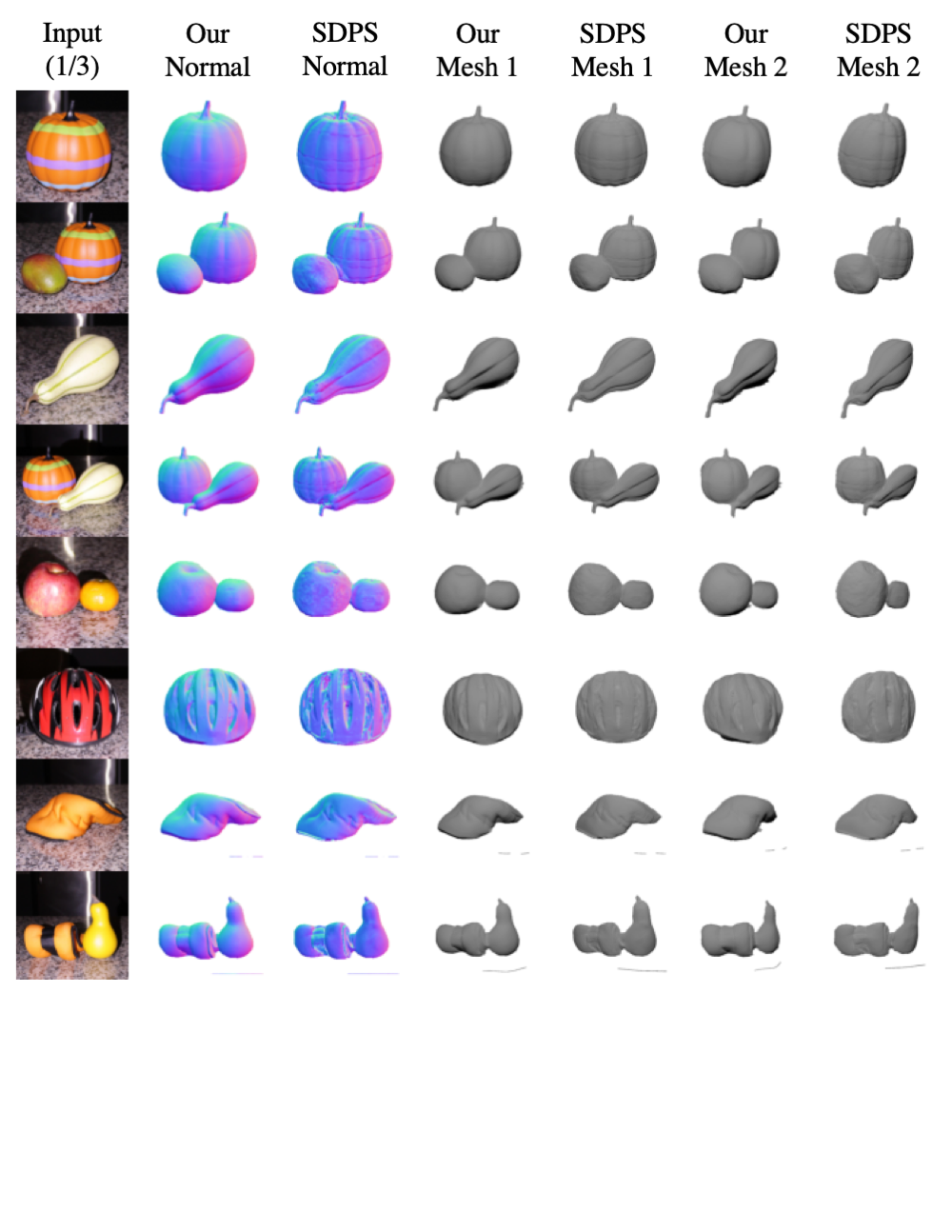}
    \vspace{-15em}
    \caption{Normal comparison with 3 input images between our method and SDPS-Net \cite{chen2019SDPS_Net} retrained with 3 input images.}
    \label{fig:multi_image_normal_b}
\end{figure*}

\begin{figure*}
    \centering
    \includegraphics[width=0.7\textwidth]{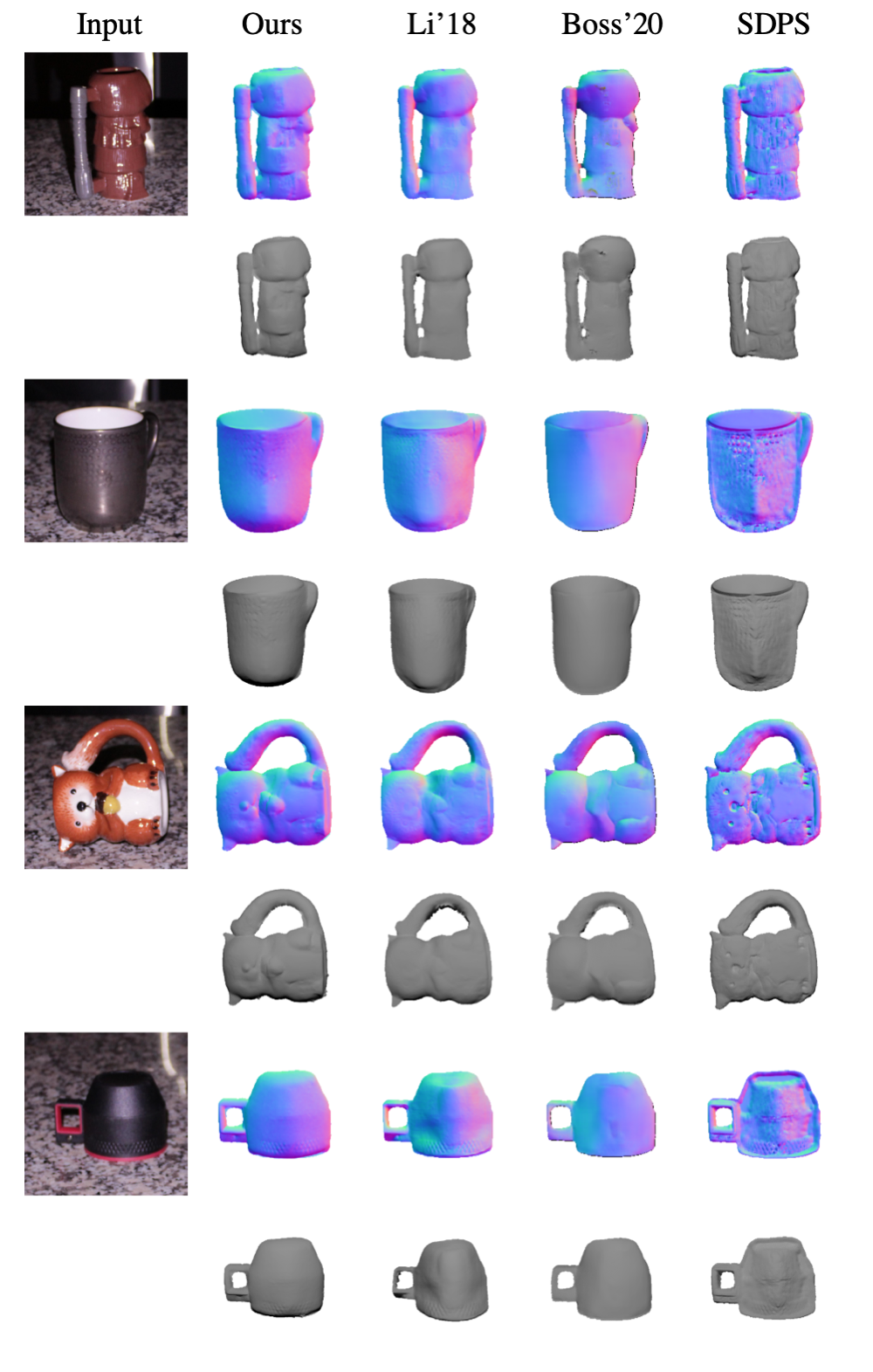}
    \caption{Normal comparison with 1 input image between our method, Li'18 \cite{li2018learning}, Boss'20 \cite{Boss2020-TwoShotShapeAndBrdf}, and SDPS-Net retrained with 1 input image \cite{chen2019SDPS_Net}}.
    \label{fig:single_image_normal_a}
\end{figure*}

\begin{figure*}
    \centering
    \includegraphics[width=0.7\textwidth]{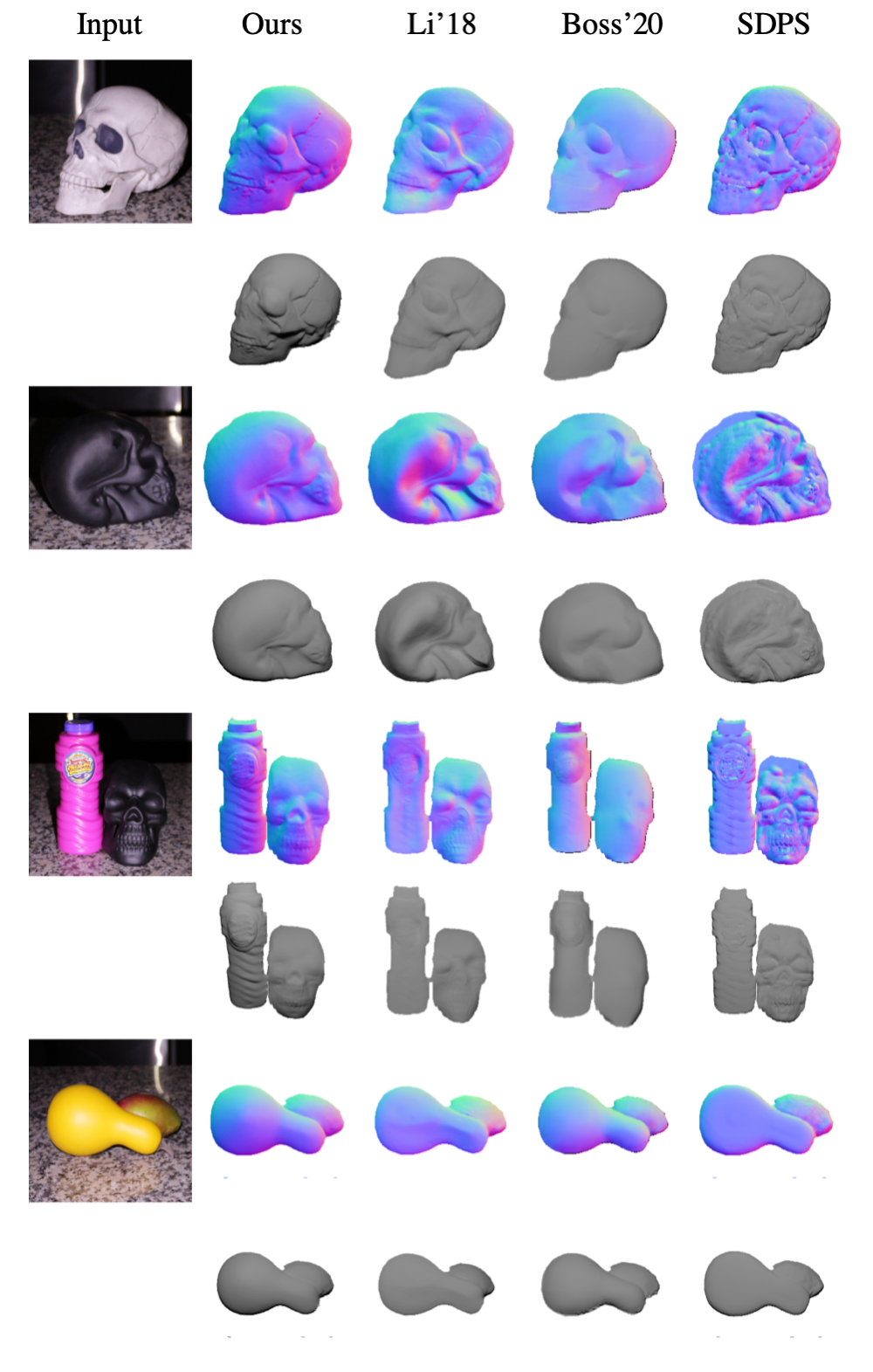}
    \caption{Normal comparison with 1 input image between our method, Li'18 \cite{li2018learning}, Boss'20 \cite{Boss2020-TwoShotShapeAndBrdf}, and SDPS-Net retrained with 1 input image \cite{chen2019SDPS_Net}}.
    \label{fig:single_image_normal_b}
\end{figure*}

\begin{figure*}
    \centering
    \includegraphics[width=0.7\textwidth]{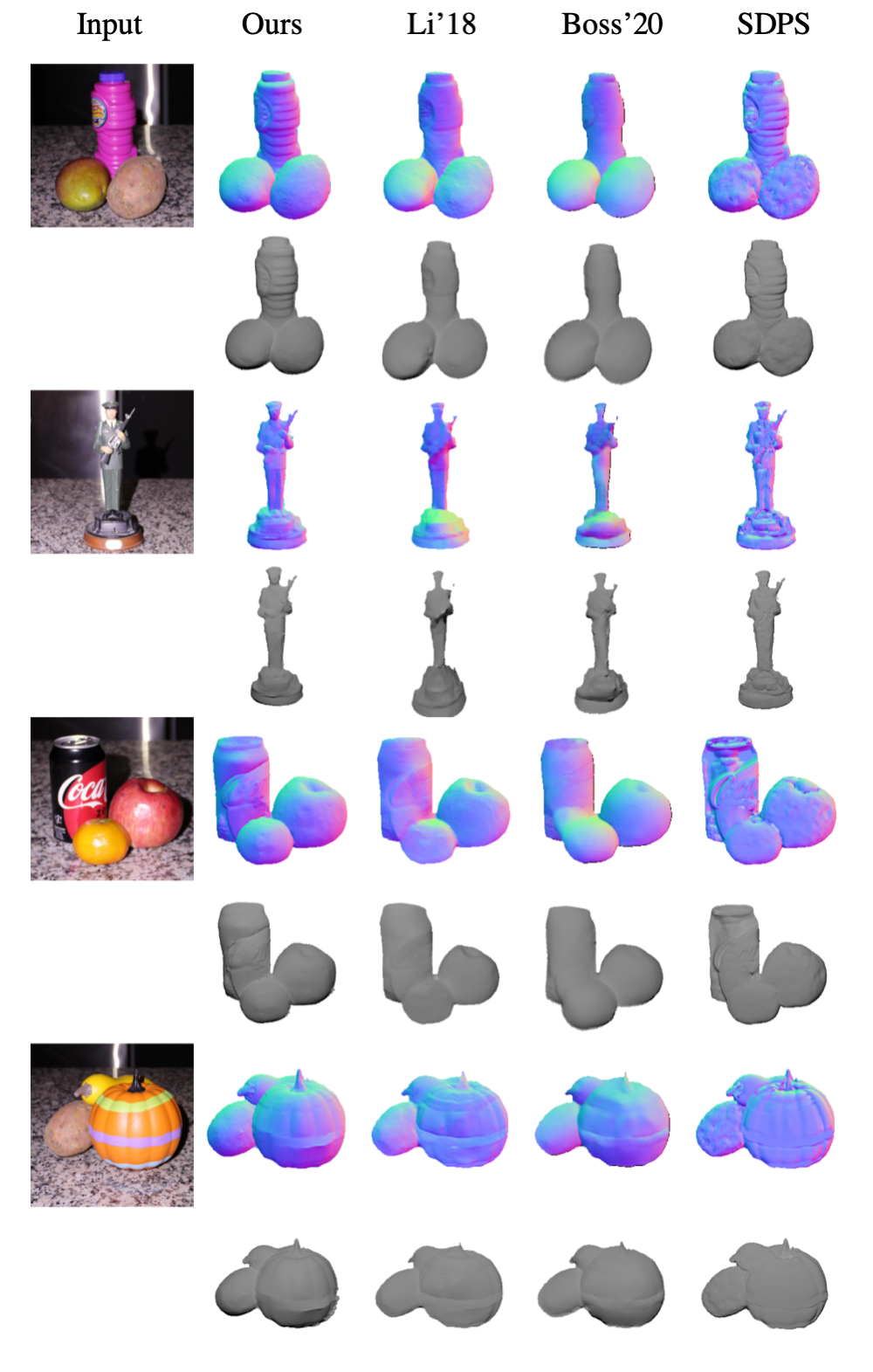}
    \caption{Normal comparison with 1 input image between our method, Li'18 \cite{li2018learning}, Boss'20 \cite{Boss2020-TwoShotShapeAndBrdf}, and SDPS-Net retrained with 1 input image \cite{chen2019SDPS_Net}.}
    \label{fig:single_image_normal_c}
\end{figure*}

\begin{figure*}
    \centering
    \includegraphics[width=0.9\textwidth]{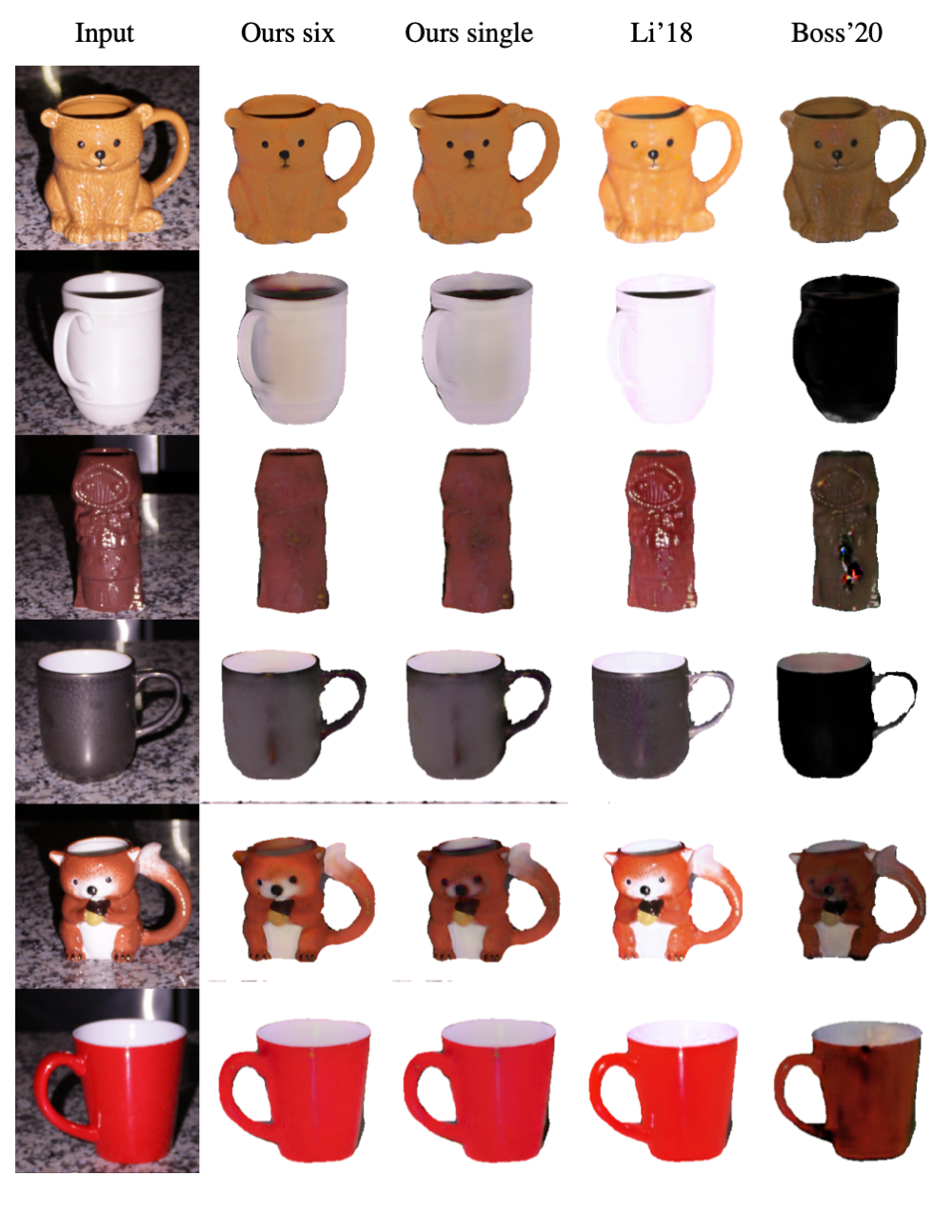}
    \vspace{-2em}
    \caption{Albedo estimation comparison between between our multi-image method with six input images (ours six), our single-image method (ours single), Li'18 \cite{li2018learning}, and Boss'20 \cite{Boss2020-TwoShotShapeAndBrdf}. Li'18 uses only a single flash image and Boss'20 uses one image with a flash and one without a flash.}
    \label{fig:single_image_albedo_a}
\end{figure*}

\begin{figure*}
    \centering
    \includegraphics[width=0.9\textwidth]{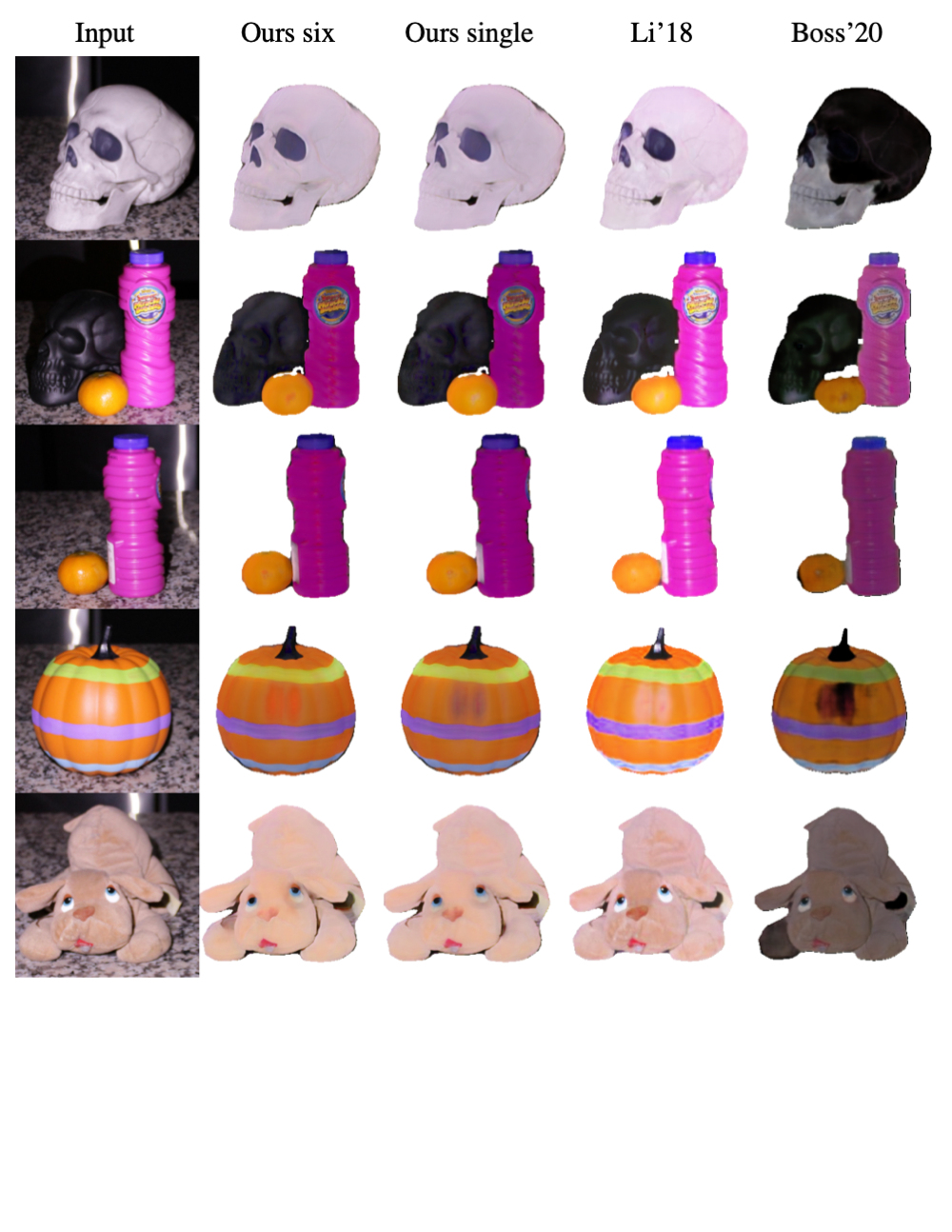}
    \vspace{-12em}
    \caption{Albedo estimation comparison between between our multi-image method with six input images (ours six), our single-image method (ours single), Li'18 \cite{li2018learning}, and Boss'20 \cite{Boss2020-TwoShotShapeAndBrdf}. Li'18 uses only a single flash image and Boss'20 uses one image with a flash and one without a flash.}
    \label{fig:single_image_albedo_b}
\end{figure*}

\begin{figure*}
    \centering
    \includegraphics[width=0.9\textwidth]{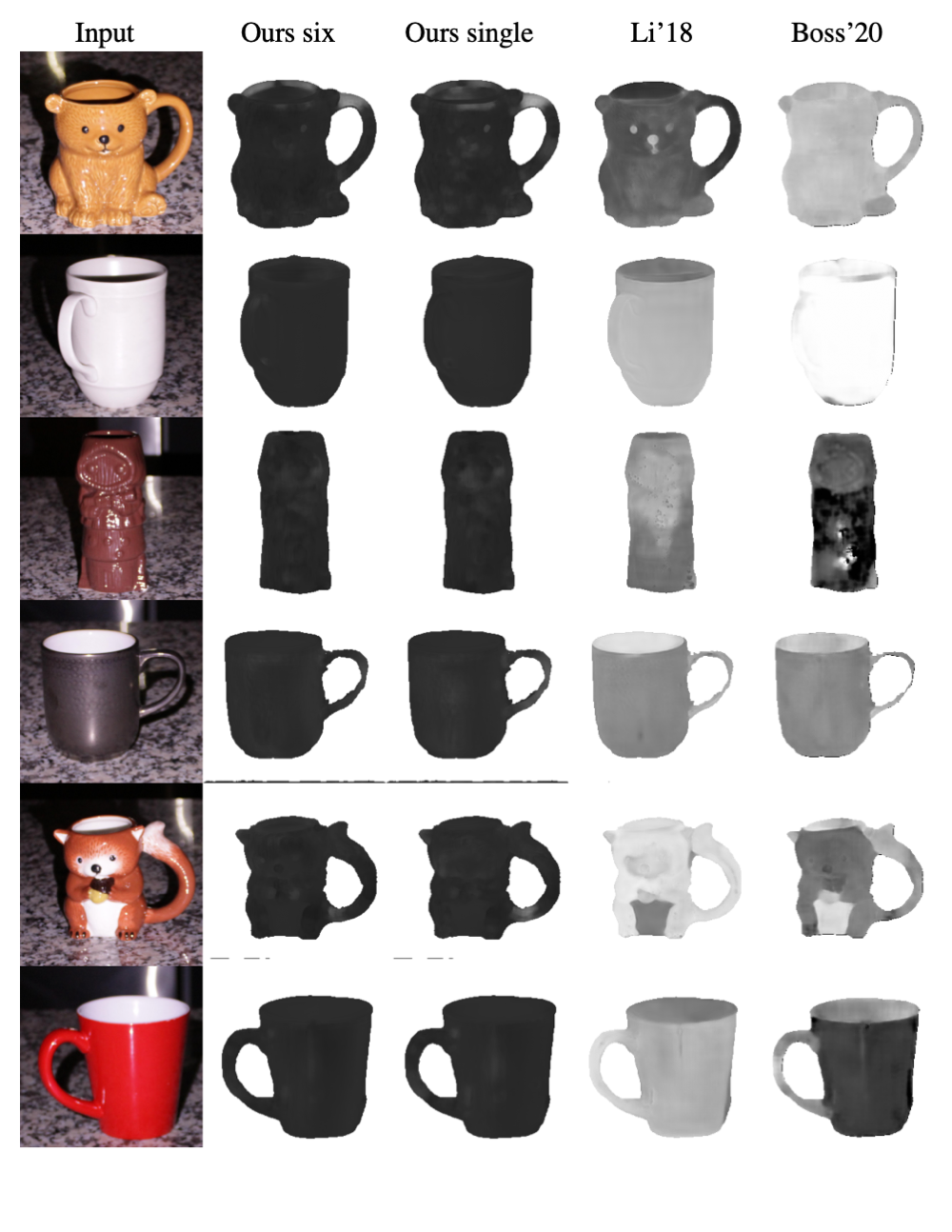}
    \vspace{-2em}
    \caption{Roughness estimation comparison between between our multi-image method with six input images (ours six), our single-image method (ours single), Li'18 \cite{li2018learning}, and Boss'20 \cite{Boss2020-TwoShotShapeAndBrdf}. Li'18 uses only a single flash image and Boss'20 uses one image with a flash and one without a flash. Brighter indicates rougher i.e. less specular.}
    \label{fig:single_image_rough_a}
\end{figure*}

\begin{figure*}
    \centering
    \includegraphics[width=0.9\textwidth]{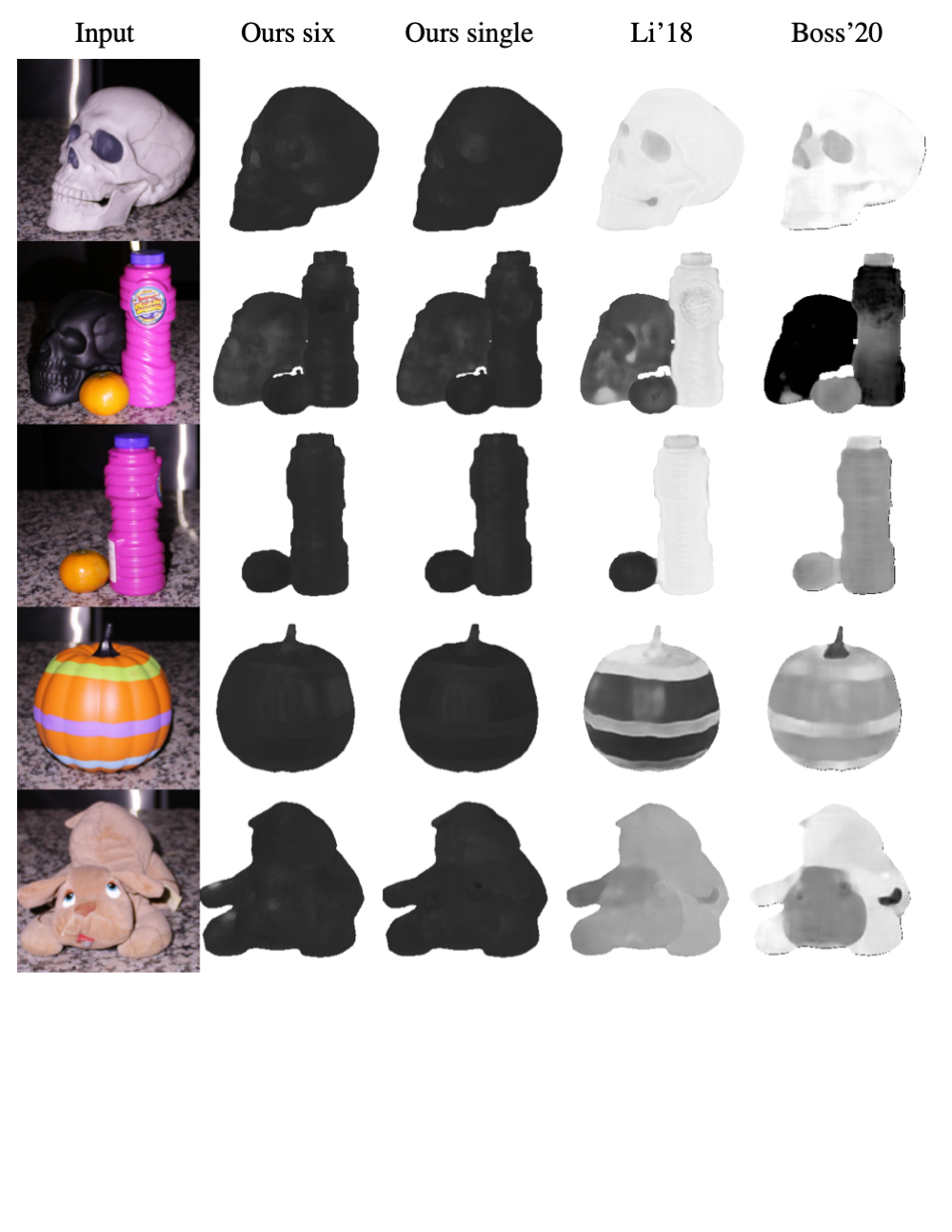}
    \vspace{-12em}
    \caption{Roughness estimation comparison between between our multi-image method with six input images (ours six), our single-image method (ours single), Li'18 \cite{li2018learning}, and Boss'20 \cite{Boss2020-TwoShotShapeAndBrdf}. Li'18 uses only a single flash image and Boss'20 uses one image with a flash and one without a flash. Brighter indicates rougher i.e. less specular.}
    \label{fig:single_image_rough_b}
\end{figure*}

Comparisons to SDPS-Net on the three image input problem are found in Figures \ref{fig:multi_image_normal_a} and \ref{fig:multi_image_normal_b}. Comparisons to Li'18 \cite{li2018learning} and Boss'20 \cite{Boss2020-TwoShotShapeAndBrdf} on the single input image problem are found in Figure \ref{fig:single_image_normal_a},\ref{fig:single_image_normal_b},\ref{fig:single_image_normal_c},\ref{fig:single_image_albedo_a}, \ref{fig:single_image_albedo_b},\ref{fig:single_image_rough_a} and \ref{fig:single_image_rough_b}.

%% file: main.bbl
\begin{thebibliography}{10}\itemsep=-1pt

\bibitem{3dtextures}
3d textures.
\newblock \url{https://3dtextures.me/}.
\newblock Accessed: 2020.

\bibitem{ackermann2015survey}
Jens Ackermann and Michael Goesele.
\newblock A survey of photometric stereo techniques.
\newblock {\em Foundations and Trends{\textregistered} in Computer Graphics and
  Vision}, 9(3-4):149--254, 2015.

\bibitem{barron2014shape}
Jonathan~T Barron and Jitendra Malik.
\newblock Shape, illumination, and reflectance from shading.
\newblock {\em IEEE transactions on pattern analysis and machine intelligence},
  37(8):1670--1687, 2014.

\bibitem{basri2007photometric}
Ronen Basri, David Jacobs, and Ira Kemelmacher.
\newblock Photometric stereo with general, unknown lighting.
\newblock {\em International Journal of computer vision}, 72(3):239--257, 2007.

\bibitem{belhumeur1999bas}
Peter~N Belhumeur, David~J Kriegman, and Alan~L Yuille.
\newblock The bas-relief ambiguity.
\newblock {\em International journal of computer vision}, 35(1):33--44, 1999.

\bibitem{bi2020deep}
Sai Bi, Zexiang Xu, Kalyan Sunkavalli, David Kriegman, and Ravi Ramamoorthi.
\newblock Deep 3d capture: Geometry and reflectance from sparse multi-view
  images.
\newblock In {\em Proceedings of the IEEE/CVF Conference on Computer Vision and
  Pattern Recognition}, pages 5960--5969, 2020.

\bibitem{Boss2020-TwoShotShapeAndBrdf}
Mark Boss, Varun Jampani, Kihwan Kim, Hendrik~P.A. Lensch, and Jan Kautz.
\newblock Two-shot spatially-varying brdf and shape estimation.
\newblock In {\em IEEE Conference on Computer Vision and Pattern Recognition
  (CVPR)}, 2020.

\bibitem{brahimi2020well}
Mohammed Brahimi, Yvain Qu{\'e}au, Bjoern Haefner, and Daniel Cremers.
\newblock On the well-posedness of uncalibrated photometric stereo under
  general lighting.
\newblock In {\em Advances in Photometric 3D-Reconstruction}, pages 147--176.
  Springer, 2020.

\bibitem{chang2015shapenet}
Angel~X Chang, Thomas Funkhouser, Leonidas Guibas, Pat Hanrahan, Qixing Huang,
  Zimo Li, Silvio Savarese, Manolis Savva, Shuran Song, Hao Su, et~al.
\newblock Shapenet: An information-rich 3d model repository.
\newblock {\em arXiv preprint arXiv:1512.03012}, 2015.

\bibitem{chen2019SDPS_Net}
Guanying Chen, Kai Han, Boxin Shi, Yasuyuki Matsushita, and Kwan-Yee~K. Wong.
\newblock Sdps-net: Self-calibrating deep photometric stereo networks.
\newblock In {\em CVPR}, 2019.

\bibitem{chen2019self}
Guanying Chen, Kai Han, Boxin Shi, Yasuyuki Matsushita, and Kwan-Yee~K Wong.
\newblock Self-calibrating deep photometric stereo networks.
\newblock In {\em Proceedings of the IEEE Conference on Computer Vision and
  Pattern Recognition}, pages 8739--8747, 2019.

\bibitem{chen2020deep}
Guanying Chen, Kai Han, Boxin Shi, Yasuyuki Matsushita, and Kwan-Yee~Kenneth
  Wong.
\newblock Deep photometric stereo for non-lambertian surfaces.
\newblock {\em IEEE Transactions on Pattern Analysis and Machine Intelligence},
  2020.

\bibitem{Chen_2018_ECCV}
Guanying Chen, Kai Han, and Kwan-Yee~K. Wong.
\newblock Ps-fcn: A flexible learning framework for photometric stereo.
\newblock In {\em Proceedings of the European Conference on Computer Vision
  (ECCV)}, September 2018.

\bibitem{chen2018ps}
Guanying Chen, Kai Han, and Kwan-Yee~K Wong.
\newblock Ps-fcn: A flexible learning framework for photometric stereo.
\newblock In {\em Proceedings of the European Conference on Computer Vision
  (ECCV)}, pages 3--18, 2018.

\bibitem{chen2020learned}
Guanying Chen, Michael Waechter, Boxin Shi, Kwan-Yee~K Wong, and Yasuyuki
  Matsushita.
\newblock What is learned in deep uncalibrated photometric stereo?
\newblock In {\em European Conference on Computer Vision}, 2020.

\bibitem{chen2017photographic}
Qifeng Chen and Vladlen Koltun.
\newblock Photographic image synthesis with cascaded refinement networks.
\newblock In {\em Proceedings of the IEEE international conference on computer
  vision}, pages 1511--1520, 2017.

\bibitem{cho2018semi}
Donghyeon Cho, Yasuyuki Matsushita, Yu-Wing Tai, and In~So Kweon.
\newblock Semi-calibrated photometric stereo.
\newblock {\em IEEE transactions on pattern analysis and machine intelligence},
  42(1):232--245, 2018.

\bibitem{cook-torrance}
R.~L. Cook and K.~E. Torrance.
\newblock A reflectance model for computer graphics.
\newblock {\em ACM Trans. Graph.}, 1(1):7–24, Jan. 1982.

\bibitem{durou2020advances}
Jean-Denis Durou, Maurizio Falcone, Yvain Qu{\'e}au, and Silvia Tozza.
\newblock Advances in photometric 3d-reconstruction, 2020.

\bibitem{multi-scale-normal-Fergus}
David Eigen, Christian Puhrsch, and Rob Fergus.
\newblock Depth map prediction from a single image using a multi-scale deep
  network.
\newblock In {\em Proceedings of the 27th International Conference on Neural
  Information Processing Systems - Volume 2}, NIPS'14, page 2366–2374,
  Cambridge, MA, USA, 2014. MIT Press.

\bibitem{favaro2012closed}
Paolo Favaro and Thoma Papadhimitri.
\newblock A closed-form solution to uncalibrated photometric stereo via diffuse
  maxima.
\newblock In {\em 2012 IEEE Conference on Computer Vision and Pattern
  Recognition}, pages 821--828. IEEE, 2012.

\bibitem{georghiades2003incorporating}
Athinodoros~S Georghiades.
\newblock Incorporating the torrance and sparrow model of reflectance in
  uncalibrated photometric stereo.
\newblock In {\em null}, page 816. Ieee, 2003.

\bibitem{georgoulis2016delight}
Stamatios Georgoulis, Konstantinos Rematas, Tobias Ritschel, Mario Fritz, Luc
  Van~Gool, and Tinne Tuytelaars.
\newblock Delight-net: Decomposing reflectance maps into specular materials and
  natural illumination.
\newblock {\em arXiv preprint arXiv:1603.08240}, 2016.

\bibitem{georgoulis2017reflectance}
Stamatios Georgoulis, Konstantinos Rematas, Tobias Ritschel, Efstratios Gavves,
  Mario Fritz, Luc Van~Gool, and Tinne Tuytelaars.
\newblock Reflectance and natural illumination from single-material specular
  objects using deep learning.
\newblock {\em IEEE transactions on pattern analysis and machine intelligence},
  40(8):1932--1947, 2017.

\bibitem{haefner2019photometric}
Bjoern Haefner, Yvain Qu{\'e}au, and Daniel Cremers.
\newblock Photometric segmentation: Simultaneous photometric stereo and
  masking.
\newblock In {\em 2019 International Conference on 3D Vision (3DV)}, pages
  222--229. IEEE, 2019.

\bibitem{haefner2019variational}
Bjoern Haefner, Zhenzhang Ye, Maolin Gao, Tao Wu, Yvain Qu{\'e}au, and Daniel
  Cremers.
\newblock Variational uncalibrated photometric stereo under general lighting.
\newblock {\em arXiv preprint arXiv:1904.03942}, 2019.

\bibitem{ikehata2018cnn}
Satoshi Ikehata.
\newblock Cnn-ps: Cnn-based photometric stereo for general non-convex surfaces.
\newblock In {\em Proceedings of the European Conference on Computer Vision
  (ECCV)}, pages 3--18, 2018.

\bibitem{isola2017image}
Phillip Isola, Jun-Yan Zhu, Tinghui Zhou, and Alexei~A Efros.
\newblock Image-to-image translation with conditional adversarial networks.
\newblock In {\em Computer Vision and Pattern Recognition (CVPR), 2017 IEEE
  Conference on}, 2017.

\bibitem{Mitsuba}
Wenzel Jakob.
\newblock Mitsuba renderer, 2010.
\newblock http://www.mitsuba-renderer.org.

\bibitem{jung2015one}
Jiyoung Jung, Joon-Young Lee, and In So~Kweon.
\newblock One-day outdoor photometric stereo via skylight estimation.
\newblock In {\em Proceedings of the IEEE Conference on Computer Vision and
  Pattern Recognition}, pages 4521--4529, 2015.

\bibitem{karras2017progressive}
Tero Karras, Timo Aila, Samuli Laine, and Jaakko Lehtinen.
\newblock Progressive growing of gans for improved quality, stability, and
  variation.
\newblock {\em arXiv preprint arXiv:1710.10196}, 2017.

\bibitem{kemelmacher20103d}
Ira Kemelmacher-Shlizerman and Ronen Basri.
\newblock 3d face reconstruction from a single image using a single reference
  face shape.
\newblock {\em IEEE transactions on pattern analysis and machine intelligence},
  33(2):394--405, 2010.

\bibitem{deeplaplacianPyramid}
W. {Lai}, J. {Huang}, N. {Ahuja}, and M. {Yang}.
\newblock Deep laplacian pyramid networks for fast and accurate
  super-resolution.
\newblock In {\em 2017 IEEE Conference on Computer Vision and Pattern
  Recognition (CVPR)}, pages 5835--5843, 2017.

\bibitem{Lattas_2020_CVPR}
Alexandros Lattas, Stylianos Moschoglou, Baris Gecer, Stylianos Ploumpis,
  Vasileios Triantafyllou, Abhijeet Ghosh, and Stefanos Zafeiriou.
\newblock Avatarme: Realistically renderable 3d facial reconstruction
  "in-the-wild".
\newblock In {\em Proceedings of the IEEE/CVF Conference on Computer Vision and
  Pattern Recognition (CVPR)}, June 2020.

\bibitem{li2019learning}
Junxuan Li, Antonio Robles-Kelly, Shaodi You, and Yasuyuki Matsushita.
\newblock Learning to minify photometric stereo.
\newblock In {\em Proceedings of the IEEE Conference on Computer Vision and
  Pattern Recognition}, pages 7568--7576, 2019.

\bibitem{li2020inverse}
Zhengqin Li, Mohammad Shafiei, Ravi Ramamoorthi, Kalyan Sunkavalli, and
  Manmohan Chandraker.
\newblock Inverse rendering for complex indoor scenes: Shape, spatially-varying
  lighting and svbrdf from a single image.
\newblock In {\em Proceedings of the IEEE/CVF Conference on Computer Vision and
  Pattern Recognition}, pages 2475--2484, 2020.

\bibitem{li2018learning}
Zhengqin Li, Zexiang Xu, Ravi Ramamoorthi, Kalyan Sunkavalli, and Manmohan
  Chandraker.
\newblock Learning to reconstruct shape and spatially-varying reflectance from
  a single image.
\newblock In {\em SIGGRAPH Asia 2018 Technical Papers}, page 269. ACM, 2018.

\bibitem{lin2019refinenet}
Guosheng Lin, Fayao Liu, Anton Milan, Chunhua Shen, and Ian Reid.
\newblock Refinenet: Multi-path refinement networks for dense prediction.
\newblock {\em IEEE Transactions on Pattern Analysis and Machine Intelligence},
  2019.

\bibitem{Lin:2017:RefineNet}
G. Lin, A. Milan, C. Shen, and I. Reid.
\newblock Refine{N}et: {M}ulti-path refinement networks for high-resolution
  semantic segmentation.
\newblock In {\em CVPR}, July 2017.

\bibitem{liu2017material}
Guilin Liu, Duygu Ceylan, Ersin Yumer, Jimei Yang, and Jyh-Ming Lien.
\newblock Material editing using a physically based rendering network.
\newblock In {\em Proceedings of the IEEE International Conference on Computer
  Vision}, pages 2261--2269, 2017.

\bibitem{logothetis2020px}
Fotios Logothetis, Ignas Budvytis, Roberto Mecca, and Roberto Cipolla.
\newblock Px-net: Simple, efficient pixel-wise training of photometric stereo
  networks.
\newblock {\em arXiv preprint arXiv:2008.04933}, 2020.

\bibitem{LN2012}
S. Lombardi and K. Nishino.
\newblock Reflectance and natural illumination from a single image.
\newblock In {\em Proceedings of European Conference on Computer Vision
  ({ECCV})}, October 2012.

\bibitem{LN2016}
S. Lombardi and K. Nishino.
\newblock Reflectance and illumination recovery in the wild.
\newblock {\em IEEE Trans. on Pattern Analysis and Machine Intelligence
  ({TPAMI})}, 38(1):129--141, January 2016.

\bibitem{lu2017symps}
Feng Lu, Xiaowu Chen, Imari Sato, and Yoichi Sato.
\newblock Symps: Brdf symmetry guided photometric stereo for shape and light
  source estimation.
\newblock {\em IEEE transactions on pattern analysis and machine intelligence},
  40(1):221--234, 2017.

\bibitem{mecca2016unifying}
Roberto Mecca and Yvain Qu{\'e}au.
\newblock Unifying diffuse and specular reflections for the photometric stereo
  problem.
\newblock In {\em 2016 IEEE Winter Conference on Applications of Computer
  Vision (WACV)}, pages 1--9. IEEE, 2016.

\bibitem{mecca2014direct}
Roberto Mecca, Ariel Tankus, Aaron Wetzler, and Alfred~M Bruckstein.
\newblock A direct differential approach to photometric stereo with perspective
  viewing.
\newblock {\em SIAM Journal on Imaging Sciences}, 7(2):579--612, 2014.

\bibitem{LIMEMeka:2018}
Abhimitra Meka, Maxim Maximov, Michael Zollhoefer, Avishek Chatterjee,
  Hans-Peter Seidel, Christian Richardt, and Christian Theobalt.
\newblock Lime: Live intrinsic material estimation.
\newblock In {\em Proceedings of Computer Vision and Pattern Recognition
  ({CVPR})}, June 2018.

\bibitem{mildenhall2020nerf}
Ben Mildenhall, Pratul~P. Srinivasan, Matthew Tancik, Jonathan~T. Barron, Ravi
  Ramamoorthi, and Ren Ng.
\newblock Nerf: Representing scenes as neural radiance fields for view
  synthesis.
\newblock In {\em ECCV}, 2020.

\bibitem{mo2018uncalibrated}
Zhipeng Mo, Boxin Shi, Feng Lu, Sai-Kit Yeung, and Yasuyuki Matsushita.
\newblock Uncalibrated photometric stereo under natural illumination.
\newblock In {\em Proceedings of the IEEE Conference on Computer Vision and
  Pattern Recognition}, pages 2936--2945, 2018.

\bibitem{Mitsuba2}
Merlin Nimier-David, Delio Vicini, Tizian Zeltner, and Wenzel Jakob.
\newblock Mitsuba 2: A retargetable forward and inverse renderer.
\newblock {\em ACM Trans. Graph.}, 38(6), Nov. 2019.

\bibitem{okabe2009attached}
Takahiro Okabe, Imari Sato, and Yoichi Sato.
\newblock Attached shadow coding: Estimating surface normals from shadows under
  unknown reflectance and lighting conditions.
\newblock In {\em 2009 IEEE 12th International Conference on Computer Vision},
  pages 1693--1700. IEEE, 2009.

\bibitem{queau2015solving}
Yvain Qu{\'e}au, Fran{\c{c}}ois Lauze, and Jean-Denis Durou.
\newblock Solving uncalibrated photometric stereo using total variation.
\newblock {\em Journal of Mathematical Imaging and Vision}, 52(1):87--107,
  2015.

\bibitem{queau2017photometric}
Yvain Qu{\'e}au, Roberto Mecca, Jean-Denis Durou, and Xavier Descombes.
\newblock Photometric stereo with only two images: A theoretical study and
  numerical resolution.
\newblock {\em Image and Vision Computing}, 57:175--191, 2017.

\bibitem{queau2017semi}
Yvain Qu{\'e}au, Tao Wu, and Daniel Cremers.
\newblock Semi-calibrated near-light photometric stereo.
\newblock In {\em International Conference on Scale Space and Variational
  Methods in Computer Vision}, pages 656--668. Springer, 2017.

\bibitem{queau2017non}
Yvain Qu{\'e}au, Tao Wu, Fran{\c{c}}ois Lauze, Jean-Denis Durou, and Daniel
  Cremers.
\newblock A non-convex variational approach to photometric stereo under
  inaccurate lighting.
\newblock In {\em Proceedings of the IEEE Conference on Computer Vision and
  Pattern Recognition}, pages 99--108, 2017.

\bibitem{rematas2016deep}
Konstantinos Rematas, Tobias Ritschel, Mario Fritz, Efstratios Gavves, and
  Tinne Tuytelaars.
\newblock Deep reflectance maps.
\newblock In {\em Proceedings of the IEEE Conference on Computer Vision and
  Pattern Recognition}, pages 4508--4516, 2016.

\bibitem{santo2017deep}
Hiroaki Santo, Masaki Samejima, Yusuke Sugano, Boxin Shi, and Yasuyuki
  Matsushita.
\newblock Deep photometric stereo network.
\newblock In {\em Proceedings of the IEEE International Conference on Computer
  Vision Workshops}, pages 501--509, 2017.

\bibitem{neuralSengupta19}
Soumyadip Sengupta, Jinwei Gu, Kihwan Kim, Guilin Liu, David~W. Jacobs, and Jan
  Kautz.
\newblock Neural inverse rendering of an indoor scene from a single image.
\newblock In {\em International Conference on Computer Vision (ICCV)}, 2019.

\bibitem{sengupta2018sfsnet}
Soumyadip Sengupta, Angjoo Kanazawa, Carlos~D Castillo, and David~W Jacobs.
\newblock Sfsnet: Learning shape, reflectance and illuminance of facesin the
  wild'.
\newblock In {\em Proceedings of the IEEE Conference on Computer Vision and
  Pattern Recognition}, pages 6296--6305, 2018.

\bibitem{sengupta2018solving}
Soumyadip Sengupta, Hao Zhou, Walter Forkel, Ronen Basri, Tom Goldstein, and
  David Jacobs.
\newblock Solving uncalibrated photometric stereo using fewer images by jointly
  optimizing low-rank matrix completion and integrability.
\newblock {\em Journal of Mathematical Imaging and Vision}, 60(4):563--575,
  2018.

\bibitem{diligent_data}
Boxin Shi, Zhe~Wu Mo, Dinglong Duan, Sai-Kit Yeung, and Ping Tan.
\newblock A benchmark dataset and evalution for non-lambertian and uncalibrated
  photometric stereo.
\newblock {\em IEEE Trans. on Pattern Analysis and Machine Intelligence
  ({TPAMI})}, 41(2):271--284, 2019.

\bibitem{fbrs2020_mask_making}
Konstantin Sofiiuk, Ilia Petrov, Olga Barinova, , and Anton Konushin.
\newblock f-brs: Rethinking backpropagating refinement for interactive
  segmentation.
\newblock {\em arXiv preprint arXiv:2001.10331}, 2020.

\bibitem{taniai2018neural}
Tatsunori Taniai and Takanori Maehara.
\newblock Neural inverse rendering for general reflectance photometric stereo.
\newblock {\em arXiv preprint arXiv:1802.10328}, 2018.

\bibitem{multi-scale-normal-gupta}
X. {Wang}, D.~F. {Fouhey}, and A. {Gupta}.
\newblock Designing deep networks for surface normal estimation.
\newblock In {\em 2015 IEEE Conference on Computer Vision and Pattern
  Recognition (CVPR)}, pages 539--547, 2015.

\bibitem{sculpture_data}
Olivia Wiles and Andrew Zisserman.
\newblock Silnet : Single- and multi-view reconstruction by learning from
  silhouettes.
\newblock In {\em British Machine Vision Conference 2017, {BMVC} 2017, London,
  UK, September 4-7, 2017}. {BMVA} Press, 2017.

\bibitem{woodham1980photometric}
Robert~J Woodham.
\newblock Photometric method for determining surface orientation from multiple
  images.
\newblock {\em Optical engineering}, 19(1):191139, 1980.

\bibitem{xu2018deep}
Zexiang Xu, Kalyan Sunkavalli, Sunil Hadap, and Ravi Ramamoorthi.
\newblock Deep image-based relighting from optimal sparse samples.
\newblock {\em ACM Transactions on Graphics (TOG)}, 37(4):126, 2018.

\bibitem{yao2018mvsnet}
Yao Yao, Zixin Luo, Shiwei Li, Tian Fang, and Long Quan.
\newblock Mvsnet: Depth inference for unstructured multi-view stereo.
\newblock {\em European Conference on Computer Vision (ECCV)}, 2018.

\bibitem{yao2019recurrent}
Yao Yao, Zixin Luo, Shiwei Li, Tianwei Shen, Tian Fang, and Long Quan.
\newblock Recurrent mvsnet for high-resolution multi-view stereo depth
  inference.
\newblock {\em Computer Vision and Pattern Recognition (CVPR)}, 2019.

\bibitem{yu2019inverserendernet}
Ye Yu and William~AP Smith.
\newblock Inverserendernet: Learning single image inverse rendering.
\newblock In {\em Proceedings of the IEEE Conference on Computer Vision and
  Pattern Recognition}, pages 3155--3164, 2019.

\bibitem{zhang1999shape}
Ruo Zhang, Ping-Sing Tsai, James~Edwin Cryer, and Mubarak Shah.
\newblock Shape-from-shading: a survey.
\newblock {\em IEEE transactions on pattern analysis and machine intelligence},
  21(8):690--706, 1999.

\bibitem{zheng2019spline}
Qian Zheng, Yiming Jia, Boxin Shi, Xudong Jiang, Ling-Yu Duan, and Alex~C Kot.
\newblock Spline-net: Sparse photometric stereo through lighting interpolation
  and normal estimation networks.
\newblock In {\em Proceedings of the IEEE International Conference on Computer
  Vision}, pages 8549--8558, 2019.

\bibitem{CycleGAN2017}
Jun-Yan Zhu, Taesung Park, Phillip Isola, and Alexei~A Efros.
\newblock Unpaired image-to-image translation using cycle-consistent
  adversarial networkss.
\newblock In {\em Computer Vision (ICCV), 2017 IEEE International Conference
  on}, 2017.

\end{thebibliography}
